%% file: main.tex
\documentclass[runningheads]{llncs}
\usepackage[T1]{fontenc}
\usepackage{graphicx}

\usepackage{hyperref}
\usepackage{color}

\usepackage{booktabs}
\usepackage{amsmath,amssymb,amsfonts}
\usepackage{subcaption}
\usepackage{multirow}
\usepackage{stfloats}
\usepackage{rotating}
\usepackage{xcolor}
\usepackage{microtype}

\usepackage[capitalize]{cleveref}
\crefname{section}{Sec.}{Secs.}
\Crefname{section}{Section}{Sections}
\crefname{table}{Tab.}{Tabs.}
\Crefname{table}{Table}{Tables}
\crefname{figure}{Fig.}{Figs.}
\Crefname{figure}{Figure}{Figures}
\crefname{equation}{Eq.}{Eqs.}
\Crefname{equation}{Equation}{Equations}

\newcommand{\etal}{\textit{et al.}}
\newcommand{\eg}{\textit{e.g.}}

\begin{document}
\title{
    PanoSAMic: Panoramic Image Segmentation from SAM Feature Encoding and Dual View Fusion
}
\titlerunning{PanoSAMic}

\author{
    Mahdi Chamseddine\inst{1,2}\orcidID{0000-0003-4119-457X} \and
    Didier Stricker\inst{1,2}\orcidID{0000-0002-5708-6023} \and
    Jason Rambach\inst{1}\orcidID{0000-0001-8122-6789}
}

\authorrunning{M. Chamseddine et al.}
\institute{
    German Research Center for Artificial Intelligence (DFKI), Kaiserslautern, Germany \and
    RPTU Kaiserslautern-Landau, Kaiserslautern, Germany \\
    \email{firstname.lastname@dfki.de}
}

\maketitle              
\input{sec/0_abstract}
\input{sec/1_introduction}

\input{sec/2_related_work}

\input{sec/3_methodology}
\input{sec/4_results}

\input{sec/6_conclusion}


%
%
%
\bibliographystyle{splncs04}
\bibliography{main}
\input{sec/X_suppl}

\end{document}

%% file: sec/0_abstract.tex
\begin{abstract}
    Existing image foundation models are not optimized for spherical images having been trained primarily on perspective images. PanoSAMic integrates the pre-trained Segment Anything (SAM) encoder to make use of its extensive training and integrate it into a semantic segmentation model for panoramic images using multiple modalities. We modify the SAM encoder to output multi-stage features and introduce a novel spatio-modal fusion module that allows the model to select the relevant modalities and best features from each modality for different areas of the input. Furthermore, our semantic decoder uses spherical attention and dual view fusion to overcome the distortions and edge discontinuity often associated with panoramic images. PanoSAMic achieves state-of-the-art (SotA) results on Stanford2D3DS for RGB, RGB-D, and RGB-D-N modalities and on Matterport3D for RGB and RGB-D modalities. \href{https://github.com/dfki-av/PanoSAMic}{https://github.com/dfki-av/PanoSAMic}.
    
    \keywords{Spherical images  \and Semantic segmentation \and Modality fusion.}
    \vspace{-1em}
\end{abstract}

%% file: sec/1_introduction.tex
\section{Introduction} \label{sec:introduction}

Spherical images offer a new approach to sensing the environment. A single compact sensor captures a full $360^\circ$ view of a scene and enables holistic understanding of the environment without the need for calibrating or aligning multiple input sources. Recent hardware developments have also gone beyond capturing only the spherical RGB data and started integrating depth information~\cite{armeni2017joint,chang2017matterport3d}.

The unique characteristics of the spherical RGB and RGB-D (RGB + Depth) sensors have sparked interest in using them in many application areas such as robotics~\cite{chaplot2020neural}, extended and augmented vision~\cite{xu2021spherical},
autonomous driving ~\cite{ma2021densepass},
and construction~\cite{kaufmann2023ontology}
among other fields.

Recent advances in deep learning algorithms and availability of massive data amounts have made it possible to train foundation models. These are models trained on vast datasets to tackle a wide range of tasks such as image and video segmentation~\cite{kirillov2023segment}
or depth estimation~\cite{yang2024depth}. 

Even though several
panoramic datasets~\cite{armeni2017joint,chang2017matterport3d,kanayama2025tof} have been published to motivate further research, existing foundation models tend to under-perform when used for spherical image processing as seen in~\Cref{fig:teaser}. Such discrepancy in performance between perspective and panoramic images is caused by the imbalance in the data used in training such foundation models.

\input{fig/fig_teaser}

Semantic segmentation is essential for scene understanding in various applications through dense pixel-level classification. Panoramic segmentation enables comprehensive scene understanding from a single frame. Existing work on panoramic image segmentation tried to solve the distortion associated with panoramic images with image projections~\cite{eder2020tangent,li2022omnifusion}, positional encoding~\cite{li2023sgat4pass}, and deformable embeddings~\cite{guttikonda2024single}.

In this work, we use the encoder of the Segment Anything Model (SAM)~\cite{kirillov2023segment}, a pioneering foundation model for image segmentation, into a panoramic image segmentation model. We integrate the pre-trained encoder, benefiting from huge resources and data used in training it, and introduce a new panoramic decoder that can handle the spherical nature of the input. Additionally, we introduced a novel fusion and refinement module to fuse multiple input modalities: RGB, Depth, and Normals.

\begin{itemize}
    \item [] Our main contributions can be summarized by:
    \item Integrating the pre-trained SAM encoder into a novel panoramic image segmentation model.
    \item Introducing the dual-view fusion for handling the spherical nature and object separation on image borders of panoramic images.
    \item Developing the Moving Convolutional Block Attention Module (MCBAM) for spatio-modal fusion.
    \item Achieving State-of-the-Art results on the panoramic Stanford2D3DS and Matterport3D public datasets.
\end{itemize}

The rest of the paper is structured as follows: \Cref{sec:related_work} presents an overview of the recent related works. \Cref{sec:methodology} goes in depth explaining our contribution and implementation details. We then present our experiments and results in~\Cref{sec:results}. \Cref{sec:ablations} validates our contributions through various ablations. Finally, we conclude with our final remarks in~\Cref{sec:conclusion}.

%% file: fig/fig_teaser.tex
\begin{figure}[t]
    \centering
    \begin{subfigure}{0.2437\linewidth}
        \includegraphics[width=\linewidth]{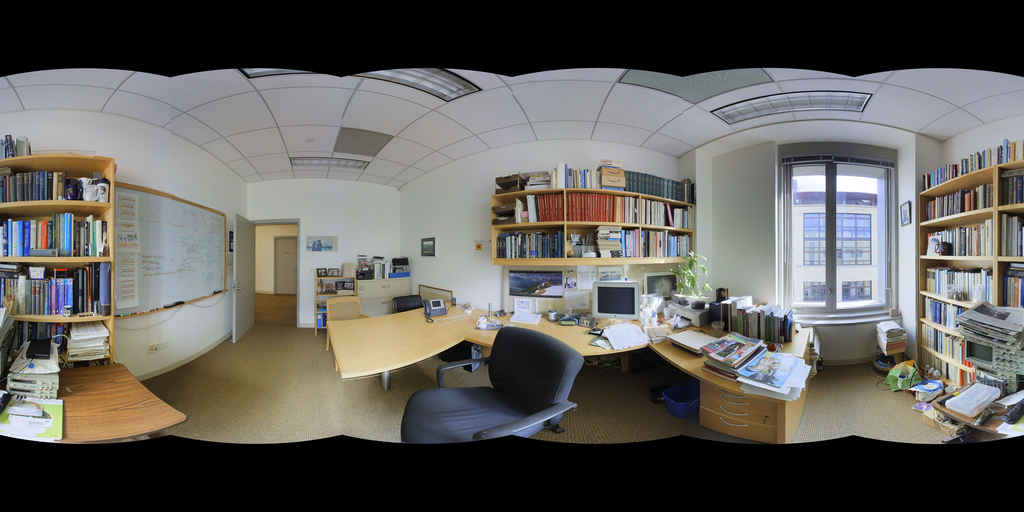}
        \vspace{-1.5em}
        \caption*{\scriptsize{RGB}}
    \end{subfigure}
    \begin{subfigure}{0.2437\linewidth}
        \includegraphics[width=\linewidth]{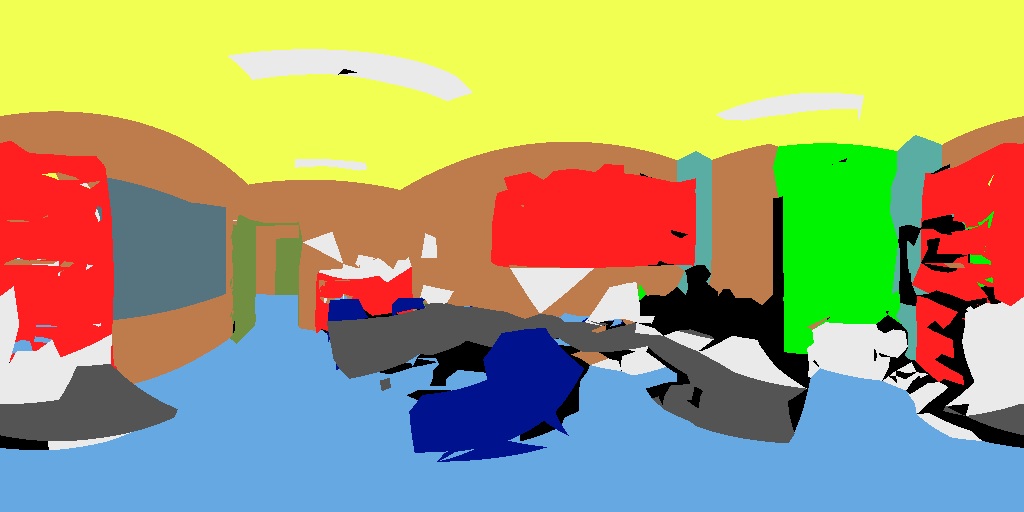}
        \vspace{-1.5em}
        \caption*{\scriptsize{Ground Truth}}
    \end{subfigure}
    \begin{subfigure}{0.2437\linewidth}
        \includegraphics[width=\linewidth]{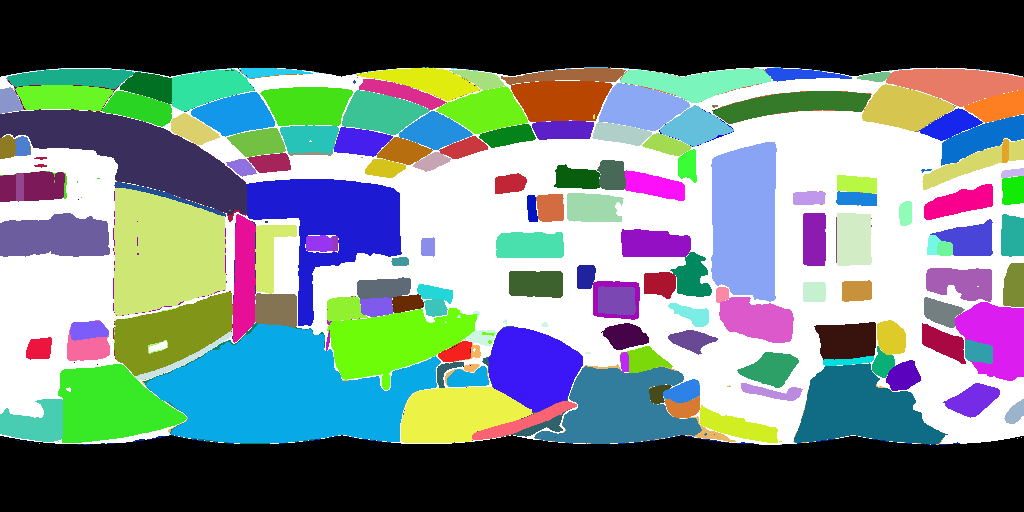}
        \vspace{-1.5em}
        \caption*{\scriptsize{SAM (instances)}}
    \end{subfigure}
    \begin{subfigure}{0.2437\linewidth}
        \includegraphics[width=\linewidth]{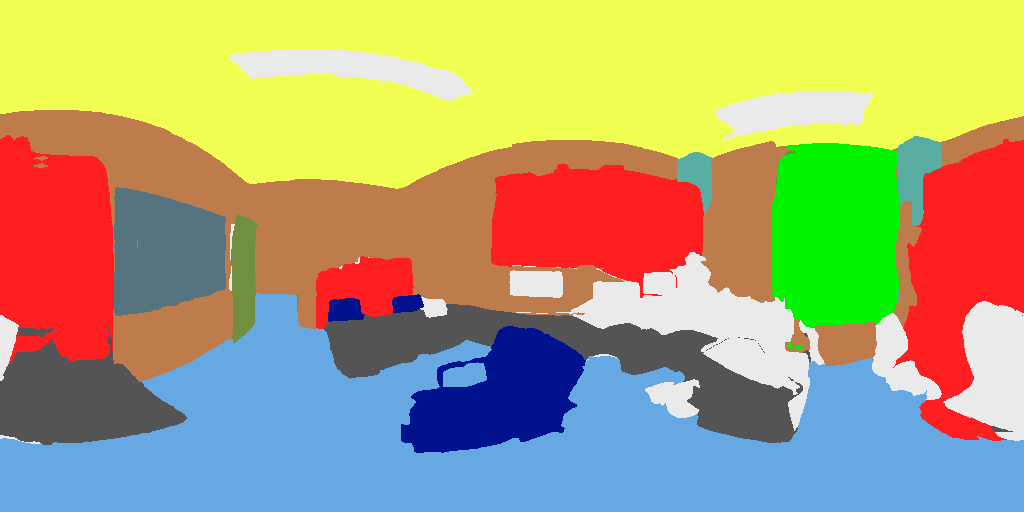}
        \vspace{-1.5em}
        \caption*{\scriptsize{PanoSAMic (semantics)}}
    \end{subfigure}
    \vspace{-0.5em}
    \caption{SAM~\cite{kirillov2023segment} is not trained for semantic segmentation and is unable to fully handle panoramic images (white is unsegmented). PanoSAMic uses the SAM pre-trained encoder and is tailored for semantic segmentation of panoramic images.}
    \label{fig:teaser}
    \vspace{-1em}
\end{figure}


%% file: sec/2_related_work.tex
\section{Related Work} \label{sec:related_work}

\subsection{Panoramic Image Segmentation}

Existing literature on panoramic and multimodal image segmentation explores techniques to handle distortions in panoramic images and the fusion of multiple modalities for improved segmentation performance.  

Early methods adapted traditional perspective-based models for wide-field-of-view images using spherical polyhedrons or tangent image representations~\cite{eder2020tangent,li2022omnifusion}. More recent techniques fall into distortion-aware and 2D geometry-aware approaches. Distortion-aware methods employ specialized convolutions, such as spherical convolutions~\cite{jiang2018spherical}, distortion-aware modules~\cite{tateno2018distortion}, and adaptive kernel fusion~\cite{zhuang2022acdnet}. Transformer-based architectures, such as Trans4PASS+~\cite{zhang2024behind},
integrate Deformable Patch Embedding and Deformable Multi-Layer Perceptron for enhanced panoramic segmentation. SGAT4PASS~\cite{li2023sgat4pass} introduces a spherical geometry-aware transformer bridging the gap between 2D panoramic segmentation and 3D-aware scene perception.

Fusion-based segmentation techniques leverage multiple data sources such as RGB, Depth, and other sensor modalities. Previous research on RGB-D segmentation developed new layers to capture geometric properties~\cite{cao2021shapeconv}
or architectures for multi-modal fusion~\cite{shen2022panoformer,sun2021hohonet}. SFSS-MMSI~\cite{guttikonda2024single} combines the deformable characteristics of Trans4PASS+ with Cross Modal Fusion (CMX)~\cite{zhang2023cmx} for panoramic semantic segmentation. 360BEV~\cite{teng2024360bev} on the other hand introduces transformer-based $360^\circ$ panorama to bird's eye view (BEV) semantic map.

\subsection{Segment Anything Model}

The Segment Anything Model (SAM)~\cite{kirillov2023segment}, is a foundation model for image segmentation trained on a large image dataset and designed to segment any object in an image.
After its introduction, SAM has been rapidly adapted to diverse domains such as medical image segmentation~\cite{ma2024segment}, 3D segmentation~\cite{yang2023sam3d}, and quality monitoring and recycling~\cite{zhou2025particlesam}
Several approaches have extended it to semantic segmentation by combining its masks with a classifier~\cite{kweon2024sam} or injecting task-specific features through adapters~\cite{yao2024sam}, 
or coupling SAM with a domain-specific encoder~\cite{zhang2023sam}.
Unlike existing work, we do not introduce additional encoders or adapters. Instead, we reuse the frozen encoder and extract and refine intermediate features.

\subsection{Open-Vocabulary Segmentation}

Recent methods have explored extending dense prediction to arbitrary categories. CAT-Seg~\cite{cho2024cat} uses CLIP~\cite{radford2021learning} to associate pixel-level features with different classes, and OpenSeeD~\cite{zhang2023simple} addresses panoptic segmentation in an open-vocabulary setting. 
Open Panoramic Segmentation (OOOPS)~\cite{zheng2024open} adapts this idea specifically to panoramic images and SAM3~\cite{carion2025sam} delivers open-vocabulary segmentation for images and videos. While these approaches highlight the promise of open-vocabulary models, supervised methods still achieve stronger results on panoramic images.


%% file: sec/3_methodology.tex
\input{fig/fig_overall_architecture}

\section{Methodology} \label{sec:methodology}


The PanoSAMic architecture, seen in~\Cref{fig:overall_architecture}, is comprised of the SAM~\cite{kirillov2023segment} model with frozen weights and modified encoder, the feature fusion module with a feature fusion block per branch output, and the semantic decoder with dual-view fusion. The network processes two views of the same panoramic scene in parallel before fusing the segmentation outputs for the final segmentation output.

\subsection{Input Modalities and Views}

To achieve better scene understanding, we leverage different modalities for their respective contributions to the segmentation task. Whereas RGB provides color and context, normal images are ideal for detecting continuous surfaces as well as edges, and depth images help mitigate the ambiguity caused by perspective distortions by introducing scale information.

While equirectangular images represent data from a $360^\circ$ scene, they introduce a discontinuity at the scene edges as shown in~\Cref{fig:shift_example} and thus reducing the segmentation quality. Existing approaches either ignore this limitation, or attempt to solve this problem through different projections~\cite{eder2020tangent,li2022omnifusion} and complex encoding~\cite{li2023sgat4pass}. Such approaches require a complicated or multi-step process for merging the projections. To address those limitations and to exploit the pre-trained SAM encoder without the need for fine-tuning, we introduce the concept of \textbf{\textit{dual-view fusion}} (shifted panoramas).

\input{fig/fig_shift_example}

PanoSAMic processes the shifted input in parallel as a new batch and the feature maps are combined in a late-fusion step after reversing the shift to produce the combined segmentation output. Formally, we describe the shifting as $X_R\left(i, j\right) = X\left(i, (j + s) \bmod W \right), \quad \forall i \in H, j \in W$,
where $X$ and $X_R$ are the original and shifted images respectively, $\left(i,j\right)$ the pixel coordinates, $H$ and $W$ the height and width of the image, and $s$ is the shift amount fixed to $W/2$. The shift ensures that an object is whole in at least one of the two views.

\subsubsection{Horizontal Positional Encoding (HPE)}

To improve the \textbf{\textit{dual-view fusion}}, we used a $1$-D positional encoding. This positional encoding is added to the encoded inputs with shifted encoding added to the shifted input. Since a horizontal shift of the equirectangular images in reality represents a rotational shift, the positional encoding allows the model to align the features for the fused segmentation.

We used the positional encoding as described in~\cite{vaswani2017attention} and defined as follows,
\begin{equation}
    \begin{split}
        &PE_{(w, 2i)} = sin(w / 10000^{2i/C}), \\
        &PE_{(w, 2i+1)} = cos(w / 10000^{2i/C}),\\
        &PE_R = PE_{(w + W/2 \bmod W, \dots),}
    \end{split}
    \label{eq:positional_encoding}
\end{equation}
where $w$ is the index of a column in the feature map, $i$ is the index of the channel, $C$ the total number of channels of the encoded features, and $W$ the total width of the encoded features. $PE_R$ is the shifted version of $PE$ produced by rolling the values of $PE$ along the width dimension. $PE$ and $PE_R$ are added to the fused features of both input views.

\subsection{Encoder Modifications}

The Segment Anything model (SAM)~\cite{kirillov2023segment} uses a plain backbone based on the vision transformer (ViT)~\cite{dosovitskiy2021an} architecture. The SAM encoder is a depth $L$ ViT with window attention at every layer except for specific layers that use global attention instead. In SAM, the number of global attention layers was chosen as $4$. The transformer blocks are followed by a convolutional block that outputs the final encoded features.

Kirillov \etal~\cite{kirillov2023segment} used three different encoder sizes with an increasing number of transformer blocks 
and subsequently  different positions of the global attention.

We extended the encoder to add a branch output after each global attention layer to allow the fusion of the features of different encoded modalities at multiple levels of encoding. This modification does not add any parameters to the encoder thus allowing for the use of the pre-trained weights.

\subsection{Feature Fusion}

The encoder processes the modalities independently as different batches and thus there is no interaction between their features. To add inter-modality feature fusion, we introduce a Fusion Block for each of the encoder branches. 

\subsection{Convolutional Block Attention Module}

The Convolutional Block Attention Module (CBAM), proposed by Woo \etal~\cite{woo2018cbam}, enhances CNN feature representation through sequential channel and spatial attention. The channel attention refines feature importance using global pooling and MLPs, while spatial attention captures spatial dependencies via pooling and convolution. CBAM is lightweight, easily integrated into CNNs, and showed performance improvements on classification and detection tasks.

\subsubsection{Moving Convolutional Block Attention Module}
\label{sec:mcbam}

In its original design, CBAM applies global pooling before attention, which works well for classification but is less suited for semantic segmentation, where different image regions often correspond to different objects.

To overcome this limitation, we introduce Moving CBAM (MCBAM), which applies a sliding-window channel attention block followed by a sliding-window spatial attention block to each of the branch outputs of the encoder. Each local region of the feature map is refined separately, allowing tailored attention in different parts of the image. When windows overlap, the channel attention values are aggregated using per channel max pooling, while spatial attention values are summed up and passed through a sigmoid activation. This overlap-handling ensures a coherent feature map across window boundaries. \Cref{fig:mcbam} shows a graphical representation of MCBAM.

The refined features are added to the input features through a feed-forward connection similar to~\cite{woo2018cbam} and then passed through a convolutional block and upscaling block.

\input{fig/fig_layers}

\subsubsection{Inter-Modality Fusion}

By applying MCBAM to the concatenated feature output of the encoder for all modalities, we allow the channel attention component to \textit{select} the best features for each window location and from the most relevant modality for that window. Similarly, the channel attention highlights the most significant areas of each window.

After inter-modality fusion, the different branch outputs are concatenated and then the HPE is added to the concatenated features before being fed into the semantic decoder.

\subsection{Semantic Decoder}
\label{sec:decoder}

For semantic segmentation, we adapt the lightweight decoder architecture of Xie \etal~\cite{xie2021segformer}, chosen for its efficiency and reliability with transformer-based (ViT~\cite{dosovitskiy2021an}) backbones. We extend it to handle the specific challenges of panoramic images. Both the original and shifted panoramas are decoded separately; the shifted view is then realigned to the original coordinate system. To merge the two predictions, we introduce a per-class blending mechanism computed with a novel spherical attention block: $x_{pred} = \alpha \cdot x_1 + (1 - \alpha)\cdot x_2$,
where $x_1$ and $x_2$ are the aligned outputs of both views, and $\alpha$ is a class-wise attention weight in $[0,1]$.

The spherical attention block, shown in~\Cref{fig:decoder_fusion}, consists of two spherical convolutional layers separated by a non-linearity and followed by a sigmoid activation. Unlike standard convolution, spherical convolution handles the wrap-around of equirectangular panoramas by padding the left border with values from the right border and vice versa, while using zero-padding at the top and bottom. This ensures that features near the left–right boundary interact consistently across the full $360^\circ$.

\subsection{Instance-Guided Semantic Refinement}

We extended SAM's instance segmentation to handle multi-modal panoramic inputs through three key modifications: (1) \textbf{Multi-modal fusion:} instance masks are generated independently from each modality and merged via greedy mask-based NMS (non-maximum suppression) to preserve complementary boundaries. (2) \textbf{Dual-view processing:} each modality is processed twice, with quality-aware NMS selecting higher-quality masks based on SAM's predicted IoU scores. (3) \textbf{Semantic refinement:} instance masks refine semantic predictions via majority voting, each instance region is assigned its most frequent semantic class, while background pixels retain original predictions.

%% file: fig/fig_overall_architecture.tex
\begin{figure}[t]
    \centering
    \includegraphics[width=\textwidth]{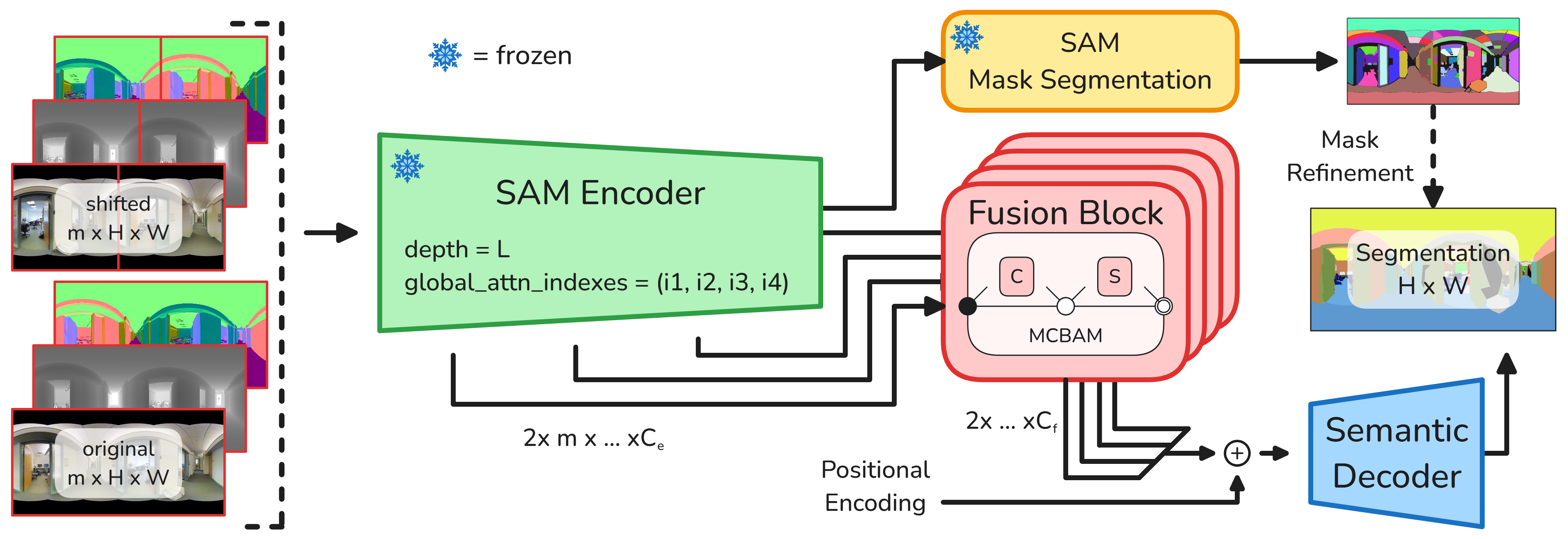}
    \caption{PanoSAMic architecture. Two views of the same panoramic input are fed into the SAM encoder. The fusion block combines and refines the features from the processed input modalities. The features are then concatenated and passed into the decoder along with a horizontal positional encoding. The semantic decoder outputs the fused segmentation which is refined with mask prediction.}
    \label{fig:overall_architecture}
    \vspace{-1.5em}
\end{figure}

%% file: fig/fig_shift_example.tex
\begin{figure}
    \centering
    \begin{subfigure}{0.45\textwidth}
        \includegraphics[width=\textwidth]{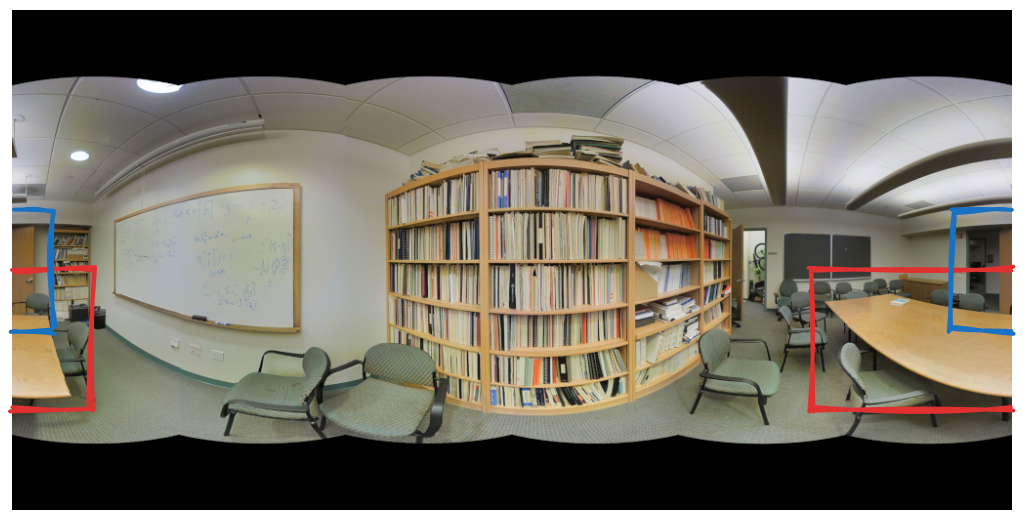}
    \end{subfigure}
    \begin{subfigure}{0.4487\textwidth}
        \includegraphics[width=\textwidth]{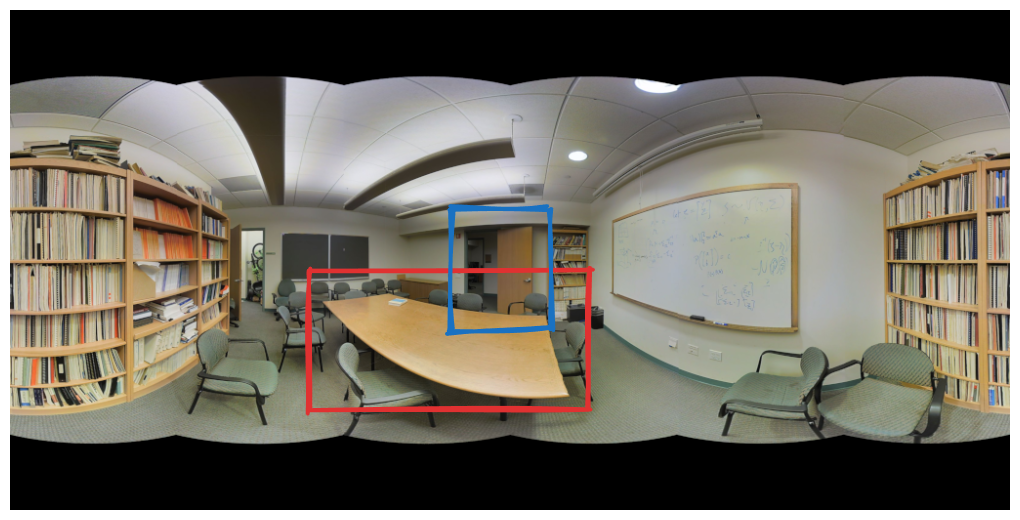}
    \end{subfigure}
    \caption{Objects in panoramic images that are disconnected on the edges are processed as whole in the shifted view.}
    \label{fig:shift_example}
    \vspace{-1.5em}
\end{figure}

%% file: fig/fig_layers.tex
\begin{figure}
    \centering
    \begin{subfigure}{0.49\textwidth}
        \includegraphics[width=\textwidth]{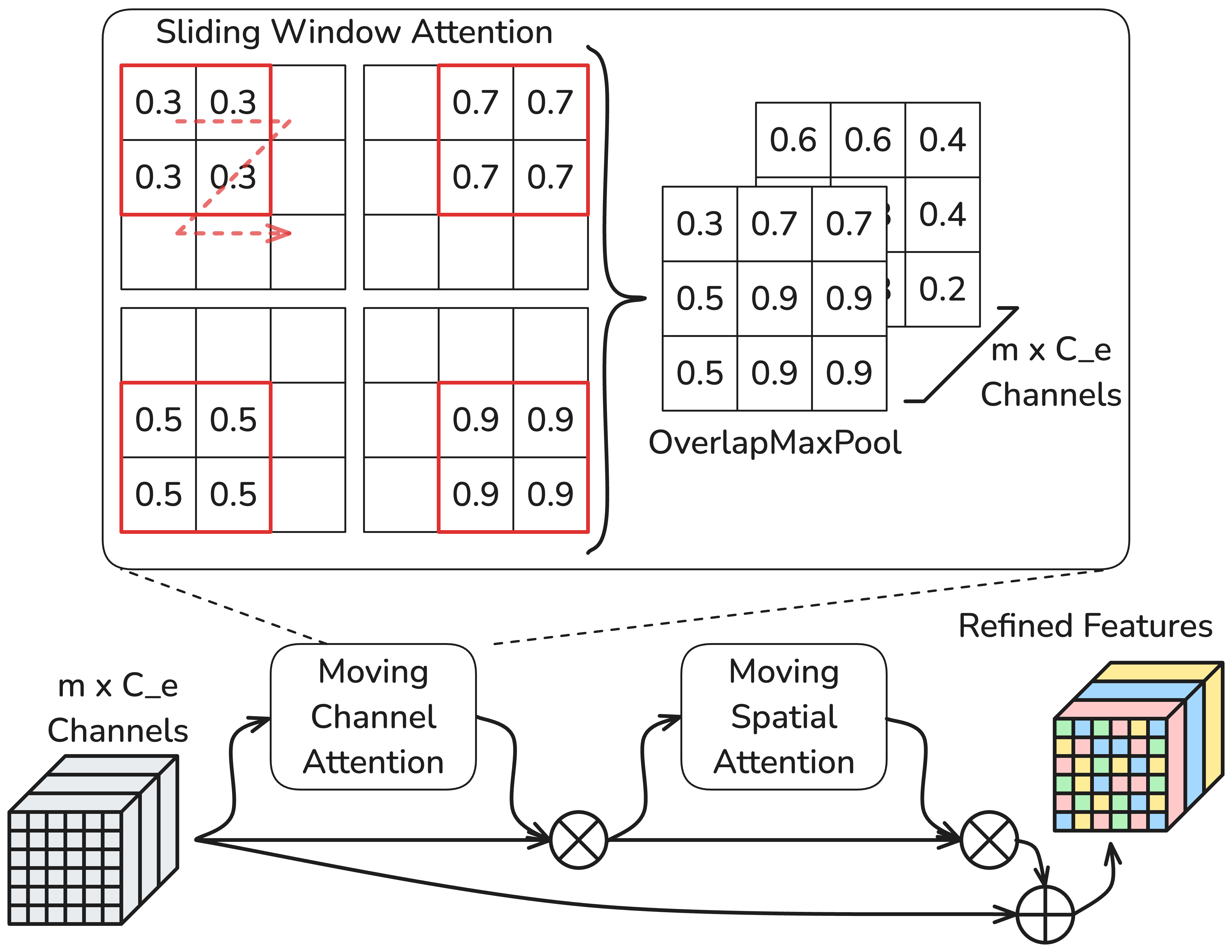}
        \caption{The MCBAM block extends the original CBAM~\cite{woo2018cbam} by applying channel and spatial attention in a sliding-window manner rather than globally.
        }
        \label{fig:mcbam}
    \end{subfigure}
    \hfill
    \begin{subfigure}{0.49\textwidth}
        \includegraphics[width=\textwidth]{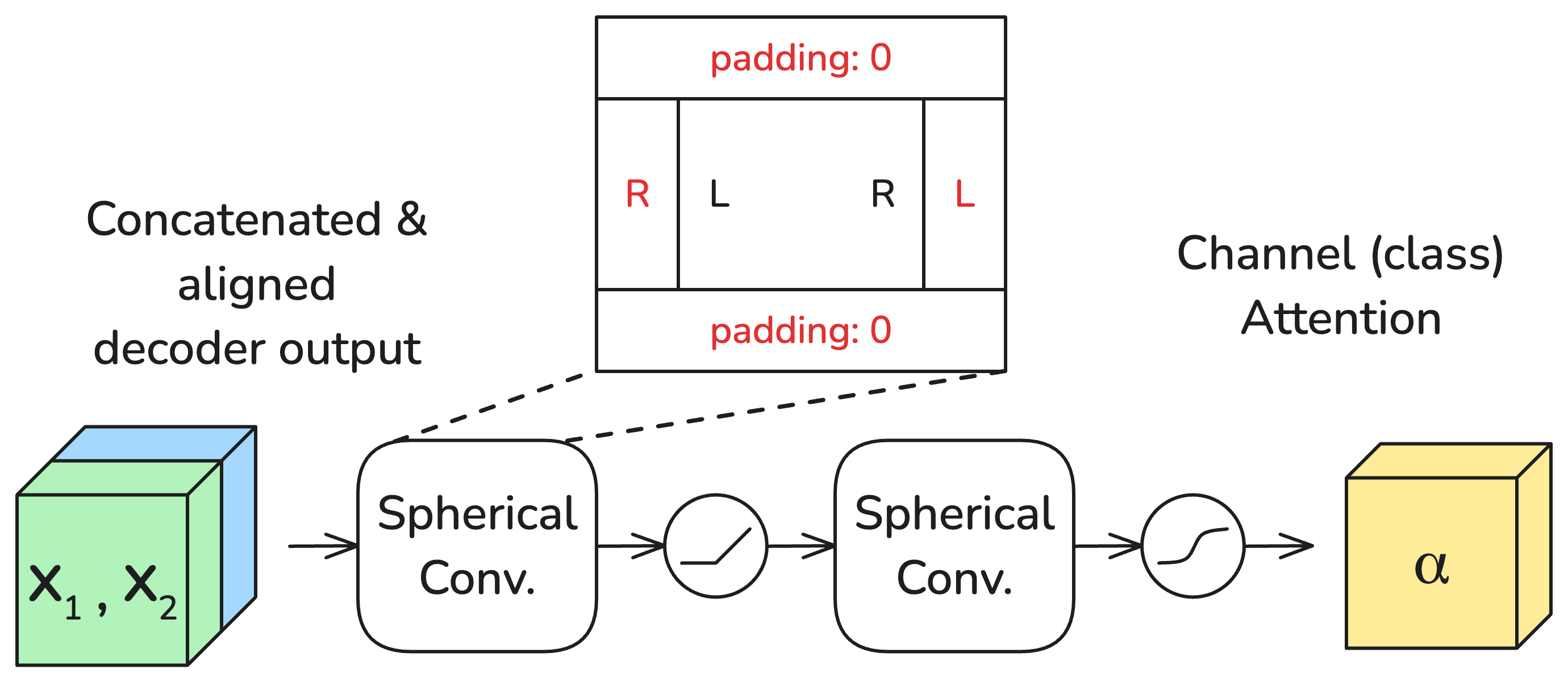}
        \caption{The spherical attention block adaptively fuses the predictions from the original and shifted panoramic views. Two spherical convolution layers compute a per-class blending weight $\alpha$, where spherical convolution handles the $360^\circ$ wrap-around by copying values from the opposite border and using zero-padding at the poles.
        }
        \label{fig:decoder_fusion}
    \end{subfigure}
    \caption{Our novel blocks used for feature fusion and dual view fusion.}
    \vspace{-1.5em}
\end{figure}

%% file: sec/4_results.tex
\section{Experiments and Results} \label{sec:results}

To evaluate the PanoSAMic architecture, we performed experiments with different modality inputs and compared to state-of-the-art methods, reporting both quantitative and qualitative results. In addition, we performed ablation experiments to evaluate the impact of specific design choices and proposed components.

\subsection{Dataset}

We used the Stanford2D3DS~\cite{armeni2017joint} and Matterport3D~\cite{chang2017matterport3d} for evaluating the model.

The \textbf{Stanford2D3DS} dataset contains $1413$ samples with RGB, depth, and normal data as well as instance and semantic labels. The data is spread over six areas and provides $13$ class labels and images at a $2048\times4096$ resolution. We used the $3$-Fold cross-validation splits suggested by the authors and evaluated our model using mean intersection over union (mIoU) and mean accuracy (mAcc).

As for the \textbf{Matterport3D} dataset, we used the pre-processed data and splits provided by Teng \etal~in BEV360~\cite{teng2024360bev} in order to perform a fair comparison. The dataset contains $10615$ samples with RGB and depth and a subset of $20$ class labels down from $40$ in the original dataset~\cite{chang2017matterport3d} to reduce the class imbalance.

\subsection{Experiment Setup}
\label{sec:experiment_setup}

For training the model, we used frozen SAM ViT-H backbone that has an encoder depth of $32$ blocks, with blocks $[8, 16, 24, 32]$ using global attention. We used a batch size of $8$ and trained for $50$ epochs, utilizing the Ranger21 optimizer~\cite{wright2021ranger21} with a maximum learning rate of $0.0005$ for Stanford2D3DS and $0.001$ for Matterport3D. Image augmentation was done using random horizontal flipping, random horizontal rolling
, and color permutation of the RGB input.

Similar to other approaches~\cite{eder2020tangent,teng2024360bev,shah2024multipanowise,guttikonda2024single}, we resized the input to $512\times1024$. We evaluated different modality inputs to the network: \textbf{RGB}, \textbf{RGB} + \textbf{D}epth, and \textbf{RGB} + \textbf{D}epth + \textbf{N}ormals. When only evaluating the \textbf{RGB} modality, and following the procedure done in other methods, we masked the black area from the metrics computation due to the lack of data in those areas. For our MCBAM (\Cref{sec:mcbam}) block we used a window size of $8\times8$ and stride of $4$ and for our $SphericalAttention$ (\Cref{sec:decoder}) we used a kernel size of $7\times7$ and a stride $1$.

For the loss, we used the Jaccard loss~\cite{rahman2016optimizing} for training on the Stanford2D3DS dataset and the alternating scheduled loss by Taubert \etal~\cite{taubert2020loss} (Cross-Entropy and Jaccard losses) for training on the Matterport3D dataset.

Since the SAM encoder was trained on images, RGB and normal maps were directly used as input to the encoder. The depth map was first preprocessed to create a pseudo-disparity image by cropping, and scaling:
\begin{equation}
    D\left(i,j\right) = 1 - \frac{\min\left(D\left(i,j\right), d_{t}\right)}{d_{t}},
\end{equation}
where $D$ is the depth map, $\left(i,j\right)$ are the pixel coordinates, and $d_{t}$ is the depth threshold representing $99.5\%$ of the depth values in the training set rounded to the nearest $10~cm$. We finally replicate $D$ to create a $3$-channel encoder input.

\subsection{Quantitative Results}

We compared the results of PanoSAMic in indoor semantic segmentation tasks and using different modality inputs with existing methods in~\Cref{tab:stanford2d3ds,tab:matterport3d}.

\subsubsection{Stanford2D3DS}

For the CBFC~\cite{zheng2023complementary}, Tangent~\cite{eder2020tangent}, 360BEV~\cite{teng2024360bev}, and MultiPanoWise~\cite{shah2024multipanowise}, we used the results reported by their respective authors in the original publications. For the rest of the methods, we used the results presented and reproduced by SFSS-MMSI~\cite{guttikonda2024single}. As for the open vocabulary methods, we evaluated the pre-trained SAM3~\cite{carion2025sam} on all classes then merged and refined the predictions 
, while for the rest of the methods we report the values from OOOPS~\cite{zheng2024open}.

\Cref{tab:stanford2d3ds} shows that we achieve state-of-the-art results across all input modalities (RGB, RGB-D, and RGB-D-N) on the Stanford2D3DS dataset. Our method consistently outperforms prior supervised approaches, like MultiPanoWise~\cite{shah2024multipanowise} and SFSS-MMSI~\cite{guttikonda2024single}, with strong gains in mean accuracy across all modalities. We also note that supervised methods surpass most open-vocabulary approaches on RGB-based semantic segmentation.

While PanoSAMic has more trainable parameters than some prior methods, the overhead is modest compared to the SAM backbone itself. Since the encoder is frozen, training remains efficient, and the extra capacity comes mainly from lightweight fusion and decoding modules. Additionally, our model size increases only marginally when moving from RGB to RGB-D and RGB-D-N ($+6$M parameters), whereas SFSS-MMSI~\cite{guttikonda2024single} grows by $+41$M between modalities.

\input{tab/tab_stanford2d3ds}





\subsubsection{Matterport3D}

We use all results as reported in the 360BEV~\cite{teng2024360bev} experiments. For the open-vocabulary methods, we report results from OOOPS~\cite{zheng2024open} and ensure fair comparison by using the same pre-processed data as 360BEV and OOOPS. SAM3~\cite{carion2025sam} results are evaluated similar to Stanford2D3DS.

On Matterport3D, \Cref{tab:matterport3d} shows overall results are lower than on Stanford2D3DS, but PanoSAMic still outperforms prior methods for RGB-only segmentation and achieves state-of-the-art performance on RGB-D, surpassing 360BEV~\cite{teng2024360bev}. As with Stanford2D3DS, supervised methods clearly outperform open-vocabulary approaches.


\input{tab/tab_matterport3d}

\subsection{Qualitative Results}

In addition to the quantitative results, we visualize our segmentation results on the Stanford2D3DS~\cite{armeni2017joint} dataset for a qualitative analysis. \Cref{fig:qualitative_results} compares the different configurations of our PanoSAMic model on different scenes.

We observe that with \textbf{RGB} input the segmentation shows better quality results on near objects than further away which appear smaller in the image. Furthermore, we see that with \textbf{RGB-D} input, more objects are segmented with finer detail, however, we notice that there was a tendency to mistake walls or doors with bookcases in some cases (\texttt{Scene 1} and \texttt{Scene 2}). Finally, and as confirmed by the quantitative results in~\Cref{tab:stanford2d3ds}, we see that the \textbf{RGB-D-N} configuration results in the highest fidelity segmentation for both near and far objects in the scenes.


Overall, the high quality of the segmentation for all tested modalities aligns with the SotA results presented in~\Cref{tab:stanford2d3ds}. 

\input{fig/fig_qualitative_results}

\subsection{Ground Truth Analysis and Generalization}

While analyzing the results of our evaluation, we observed more closely some of the worse performing samples (low mIoU) on the validation set of the different folds. This resulted in finding some samples in the dataset that had various levels of ground truth imprecision. \Cref{fig:generalization} shows different examples of these imprecisions as well as their respective segmentation results.

Such imprecisions can be caused by, quantization in mesh reprojection (Stanford2D3DS labels are projected from 3D to 2D), mislabeled instances, or missing annotations as can be seen in the outdoor scene in~\Cref{subfig:high_dist}.

Comparing the images in~\Cref{fig:generalization} to the segmentation output shows that PanoSAMic segmented scenes well and often produced better labels than the provided ground truth. Given that Stanford2D3DS is primarily an indoor dataset, the segmentation result in~\Cref{subfig:high_dist} clearly demonstrates the generalization capability of the model. The model, however, mislabeled the sky, which is expected given the absence of this class in training and the visual similarity between an overcast sky and a white wall.

\input{fig/fig_generalization}

\subsection{Ablation Studies} \label{sec:ablations}

\input{tab/tab_ablation_study}

We validate our contributions through ablations on different model configurations. We follow the setup in~\Cref{sec:experiment_setup} using \textbf{RGB-D-N} input on the standard Stanford2D3DS $1$-Fold (\texttt{Area 5}) validation. As baseline, we use the vanilla SAM encoder with a convolutional decoder with a \textbf{single view} input.

The different configurations evaluated in~\Cref{tab:ablation_study} explore alternative feature refinement strategies within a \textbf{single-view} segmentation setting. Introducing the encoder modification alone already yields a substantial improvement over the baseline, resulting in a large gain in mIoU and placing the model firmly within state-of-the-art performance for Stanford2D3DS~\cite{armeni2017joint}.

As shown in~\Cref{tab:ablation_study}, adding encoder branches without attention (B) yields a substantial improvement over the baseline configuration (A). Introducing plain Channel Attention on top of this setup (C) leads to a slight performance drop compared to no attention, indicating that channel-wise reweighting alone is not sufficient for dense semantic segmentation. Similarly, applying the standard CBAM module (D) does not consistently improve performance, suggesting limitations of its global spatial attention when operating on fine-grained, densely labeled scenes. In contrast, the proposed Moving CBAM (MCBAM) (E) recovers and further improves the mIoU by enabling localized, region-aware feature refinement. Finally, incorporating the instance refinement strategy (F) yields the best overall performance, demonstrating its complementary benefit on top of MCBAM.

To further assess the impact of dual-view fusion, we evaluated segmentation performance specifically on edge regions of the images using the $3$-Fold \textbf{RGB} validation of Stanford2D3DS. \Cref{tab:edge_ablation} reports results for different edge ratios, defined as the fraction of pixels near the left and right borders (\eg~an edge ratio of $0.5$ corresponds to $0.25W$ on each side, where $W$ is the image width).


For single-view segmentation, performance decreases as the edge ratio narrows. In contrast, dual-view fusion shows the opposite trend. The mIoU difference between single and dual view grows from $1.67\%$ for full images to $4.87\%$ at the edges, confirming that dual-view fusion effectively mitigates boundary discontinuities.


\input{tab/tab_edge_ablation}

%% file: tab/tab_stanford2d3ds.tex
\begin{table}[t]
    \centering
    \caption{Semantic segmentation results on the Stanford2D3DS~\cite{armeni2017joint} dataset using different input modalities and configurations (open vocabulary vs. supervised).
    }
    \label{tab:stanford2d3ds}
    \scriptsize
    \setlength{\tabcolsep}{6pt}
    \begin{tabular}{@{}lcccc@{}} \toprule
        \multirow{2}{*}{\textbf{Method}}              & \textbf{Configuration}          & \multicolumn{2}{c}{\textbf{3-fold Validation}} & \textbf{\# Params} \\
                                                      & \textbf{/ Modalities}           & \textbf{mIoU \%}    & \textbf{mAcc \%} & \textbf{(millions)} \\ \midrule \midrule
                                                      
        CAT-seg~\cite{cho2024cat}                     & \multirow{4}{*}{\begin{tabular}{c}Open-vocabulary\\with\\RGB\end{tabular}} & $39.60$ & -- & $59.5$ \\
        OpenSeeD~\cite{zhang2023simple}               &                                 & $40.00$             & --                  & $65.4$ \\
        OOOPS~\cite{zheng2024open}                    &                                 & $42.60$             & --                  & $8.7$ \\
        SAM3~\cite{carion2025sam}                     &                                 & $52.05$             & $65.02$             & $850$ \\ \midrule
        
        SFSS-MMSI~\cite{guttikonda2024single}         & \multirow{8}{*}{RGB}            & $52.87$             & $63.96$             & $40$ \\
        HoHoNet~\cite{sun2021hohonet}                 &                                 & $51.99$             & $62.97$             & $70$ \\
        PanoFormer~\cite{shen2022panoformer}          &                                 & $52.35$             & $64.31$             & $20$ \\
        CBFC~\cite{zheng2023complementary}            &                                 & $52.20$             & $\underline{65.60}$ & -- \\
        Tangent~\cite{eder2020tangent}                &                                 & $45.60$             & $65.20$             & -- \\
        Trans4PASS+~\cite{zhang2024behind}            &                                 & $52.04$             & $63.98$             & $39$ \\
        MultiPanoWise~\cite{shah2024multipanowise}    &                                 & $\underline{54.60}$ & --                  & -- \\
        \textbf{\textit{PanoSAMic (ours)}}            &                                 & $\mathbf{59.62}$    & $\mathbf{74.11}$    & $178$ \\ \midrule
        
        SFSS-MMSI~\cite{guttikonda2024single}         & \multirow{7}{*}{RGB-D}          & $55.49$             & $68.57$             & $81$ \\
        HoHoNet~\cite{sun2021hohonet}                 &                                 & $56.73$             & $68.23$             & $70$ \\
        PanoFormer~\cite{shen2022panoformer}          &                                 & $\underline{57.03}$ & $68.08$             & $20$ \\
        CBFC~\cite{zheng2023complementary}            &                                 & $56.70$             & $\underline{70.80}$ & -- \\
        Tangent~\cite{eder2020tangent}                &                                 & $52.50$             & $70.10$             & -- \\
        360BEV~\cite{teng2024360bev}                  &                                 & $54.30$             & --                  & $27.7$ \\
        \textbf{\textit{PanoSAMic (ours)}}            &                                 & $\mathbf{60.90}$    & $\mathbf{73.95}$    & $184$ \\ \midrule
        
        SFSS-MMSI~\cite{guttikonda2024single}         & \multirow{2}{*}{RGB-D-N}        & $\underline{59.43}$ & $\underline{69.03}$ & $123$ \\
        \textbf{\textit{PanoSAMic (ours)}}            &                                 & $\mathbf{61.57}$    & $\mathbf{74.04}$    & $191$ \\ \bottomrule
    \end{tabular}
    \vspace{-2em}
\end{table}

%% file: tab/tab_matterport3d.tex
\begin{table}[t]
    \centering
    \caption{Semantic segmentation results on the Matterport3D~\cite{chang2017matterport3d} dataset using different input modalities.}
    \label{tab:matterport3d}
    \scriptsize
    \setlength{\tabcolsep}{6pt}
    \begin{tabular}{@{}lccc@{}} \toprule
        \multirow{2}{*}{\textbf{Method}}              & \textbf{Configuration /}        & \multirow{2}{*}{\textbf{mIoU \%}} \\
                                                      & \textbf{Modalities}             &                                   \\ \midrule \midrule
                                                      
        CAT-seg~\cite{cho2024cat}                     & \multirow{4}{*}{\begin{tabular}{c}Open-vocabulary\\with\\RGB\end{tabular}}& $31.10$ \\
        OpenSeeD~\cite{zhang2023simple}               &                                 & $31.60$ \\
        OOOPS~\cite{zheng2024open}                    &                                 & $32.50$ \\
        SAM3~\cite{carion2025sam}                     &                                 & $42.29$ \\ \midrule
        
        Trans4PASS+~\cite{zhang2024behind}            & \multirow{4}{*}{RGB}            & $42.60$ \\
        HoHoNet~\cite{sun2021hohonet}                 &                                 & $44.10$ \\
        SegFormer~\cite{xie2021segformer}             &                                 & $\underline{45.53}$ \\
        \textbf{\textit{PanoSAMic (ours)}}            &                                 & $\mathbf{46.59}$ \\ \midrule
        
        360BEV~\cite{teng2024360bev}                  & \multirow{2}{*}{RGB-D}          & $\underline{46.35}$ \\
        \textbf{\textit{PanoSAMic (ours)}}            &                                 & $\mathbf{48.43}$ \\ \bottomrule
    \end{tabular}
    \vspace{-2em}
\end{table}

%% file: fig/fig_qualitative_results.tex
\begin{figure}[t]
    \centering
    \begin{subfigure}{\textwidth}
        \centering
        \rotatebox[origin=l]{90}{\begin{minipage}{4pc}\caption*{\textbf{\scriptsize{RGB}}}\end{minipage}}%
        \begin{subfigure}{0.311\textwidth}
            \includegraphics[width=\textwidth]{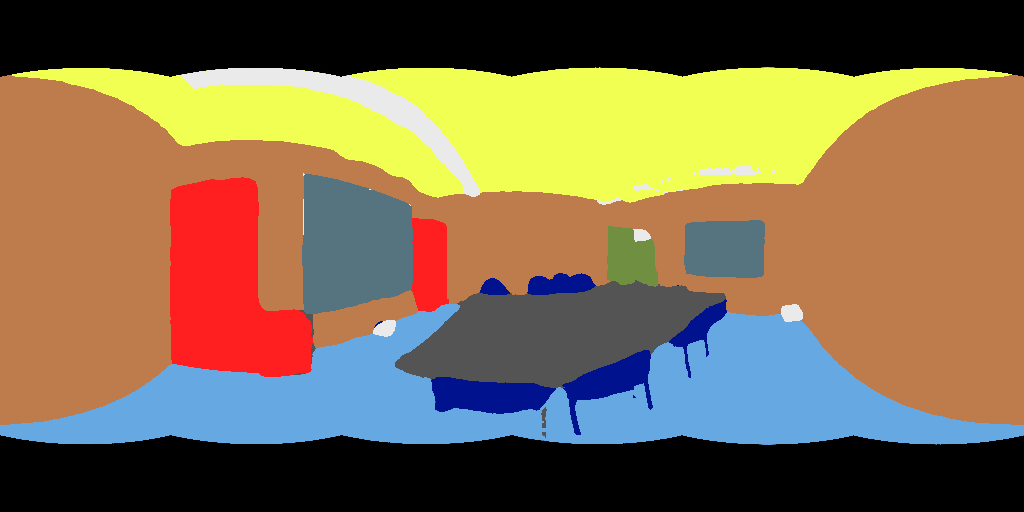}
        \end{subfigure}
        \begin{subfigure}{0.311\textwidth}
            \includegraphics[width=\textwidth]{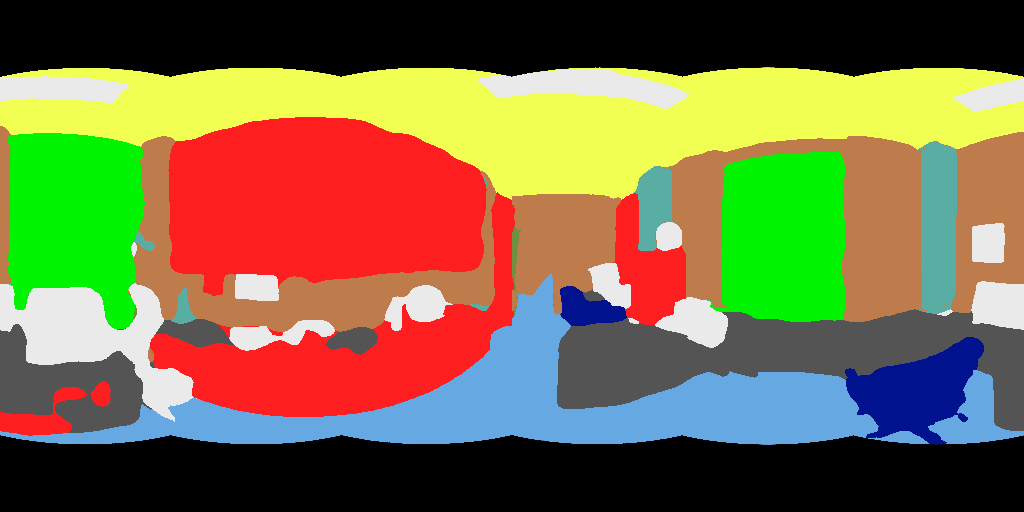}
        \end{subfigure}
        \begin{subfigure}{0.311\textwidth}
            \includegraphics[width=\textwidth]{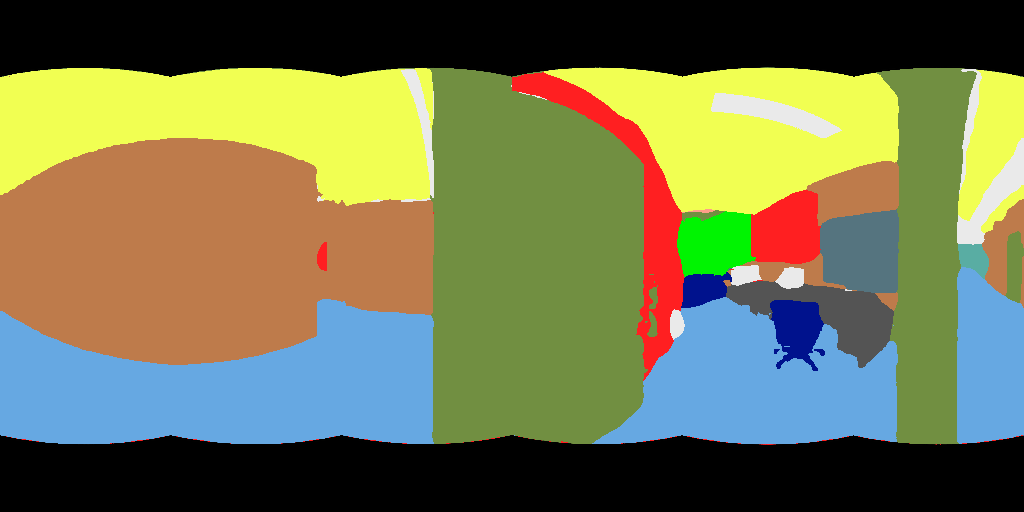}
        \end{subfigure}
    \end{subfigure}
    \begin{subfigure}{\textwidth}
        \centering
        \rotatebox[origin=l]{90}{\begin{minipage}{4pc}\caption*{\textbf{\scriptsize{RGB-D}}}\end{minipage}}%
        \begin{subfigure}{0.311\textwidth}
            \includegraphics[width=\textwidth]{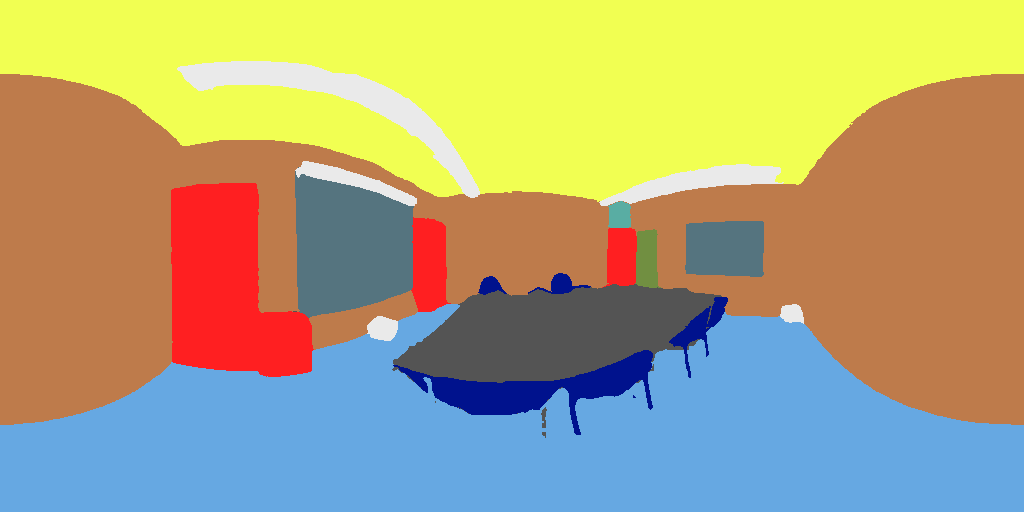}
        \end{subfigure}
        \begin{subfigure}{0.311\textwidth}
            \includegraphics[width=\textwidth]{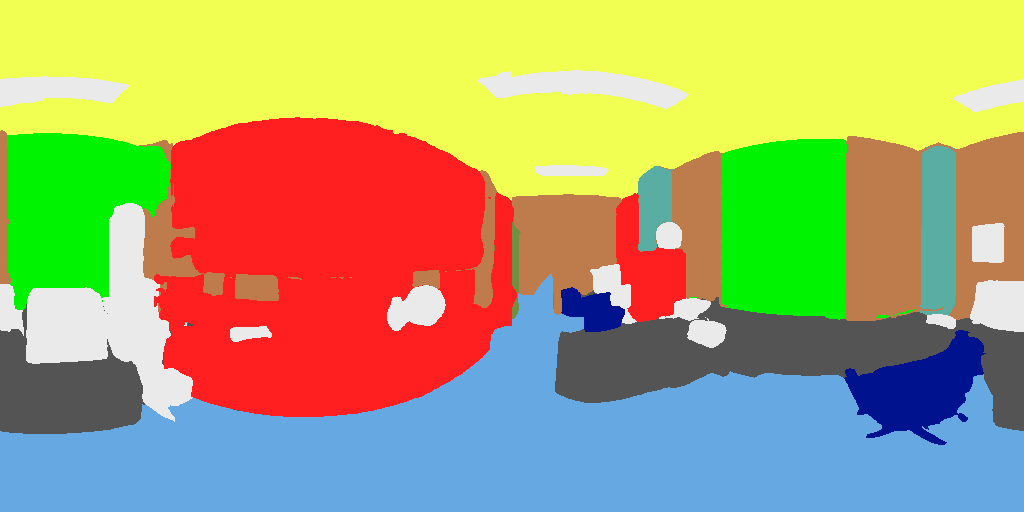}
        \end{subfigure}
        \begin{subfigure}{0.311\textwidth}
            \includegraphics[width=\textwidth]{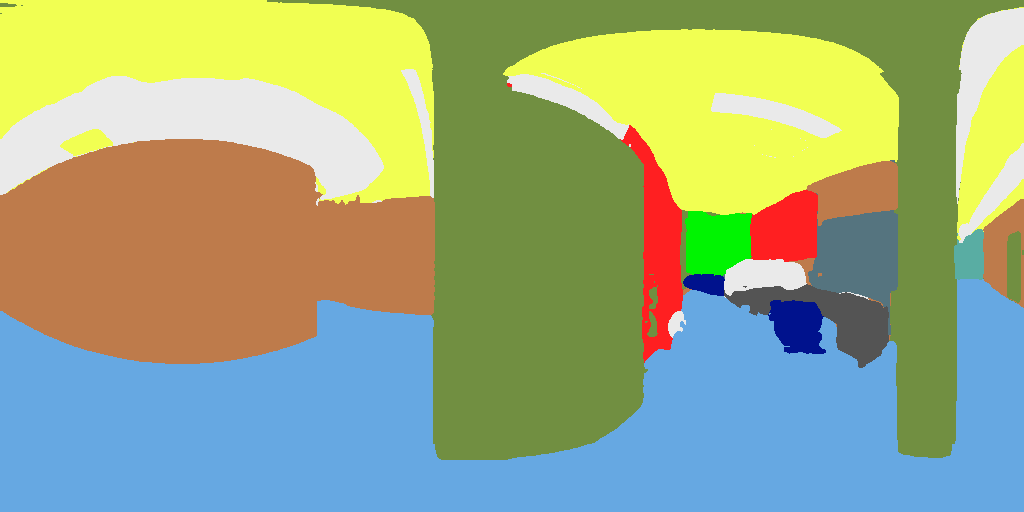}
        \end{subfigure}
    \end{subfigure}
    \begin{subfigure}{\textwidth}
        \centering
        \rotatebox[origin=l]{90}{\begin{minipage}{4pc}\caption*{\textbf{\scriptsize{RGB-D-N}}}\end{minipage}}%
        \begin{subfigure}{0.311\textwidth}
            \includegraphics[width=\textwidth]{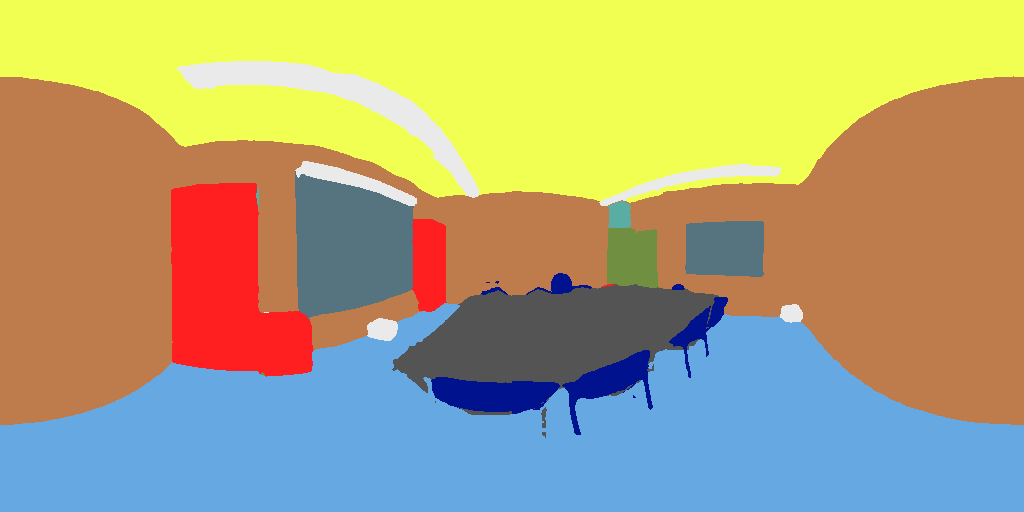}
        \end{subfigure}
        \begin{subfigure}{0.311\textwidth}
            \includegraphics[width=\textwidth]{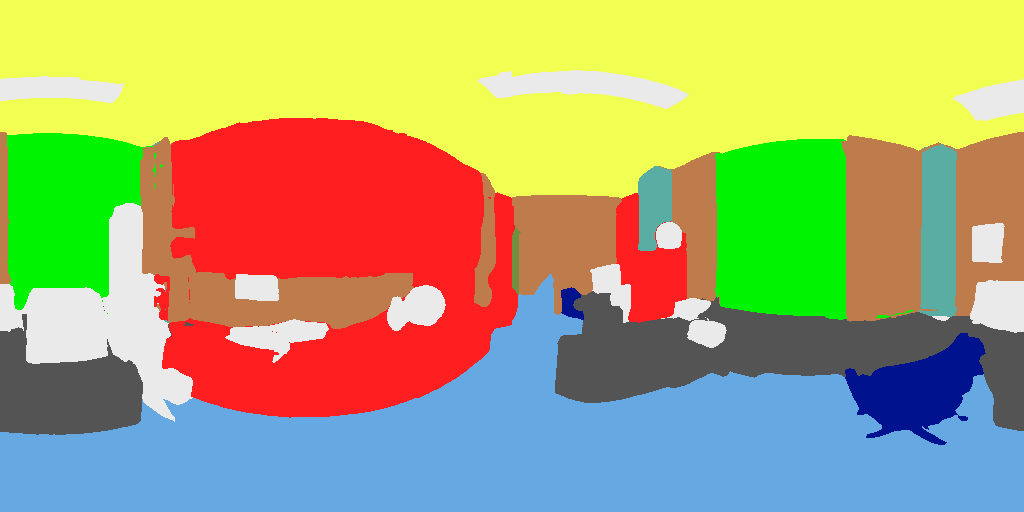}
        \end{subfigure}
        \begin{subfigure}{0.311\textwidth}
            \includegraphics[width=\textwidth]{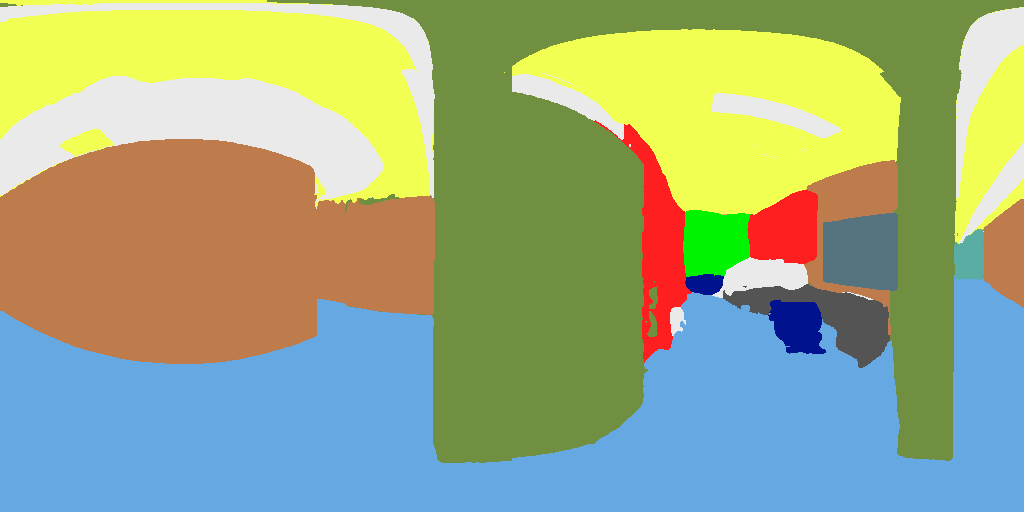}
        \end{subfigure}
    \end{subfigure}
    \begin{subfigure}{\textwidth}
        \centering
        \rotatebox[origin=l]{90}{\begin{minipage}{4pc}\caption*{\textbf{\scriptsize{GT}}}\end{minipage}}%
        \begin{subfigure}{0.311\textwidth}
            \includegraphics[width=\textwidth]{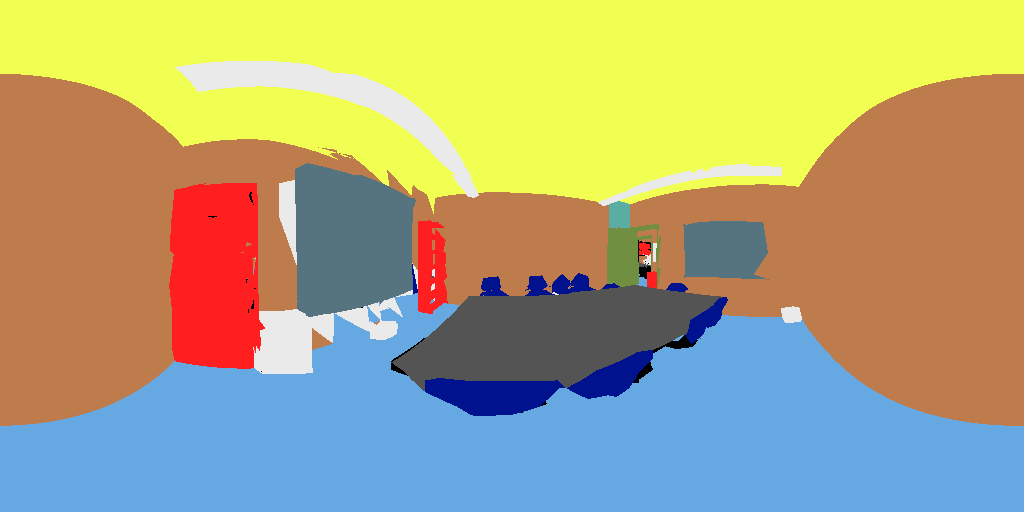}
        \end{subfigure}
        \begin{subfigure}{0.311\textwidth}
            \includegraphics[width=\textwidth]{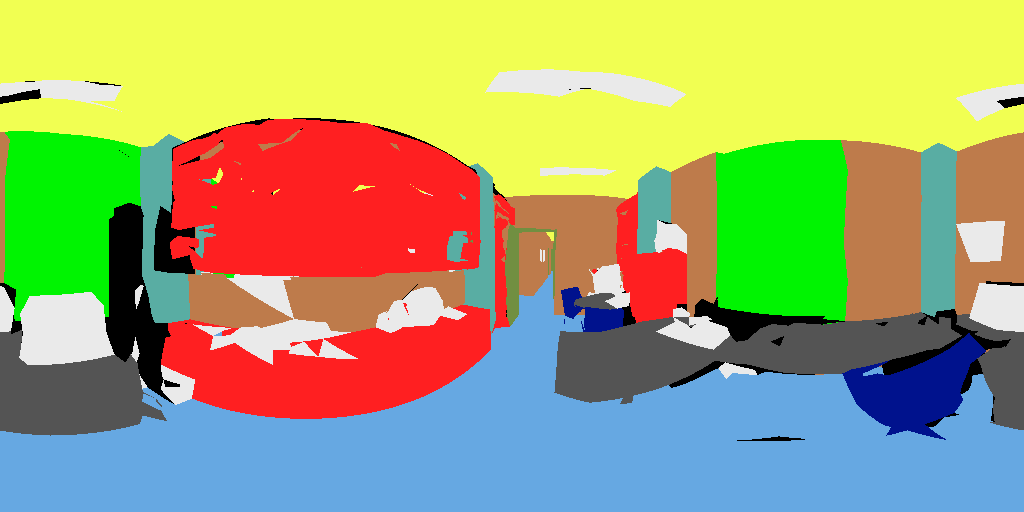}
        \end{subfigure}
        \begin{subfigure}{0.311\textwidth}
            \includegraphics[width=\textwidth]{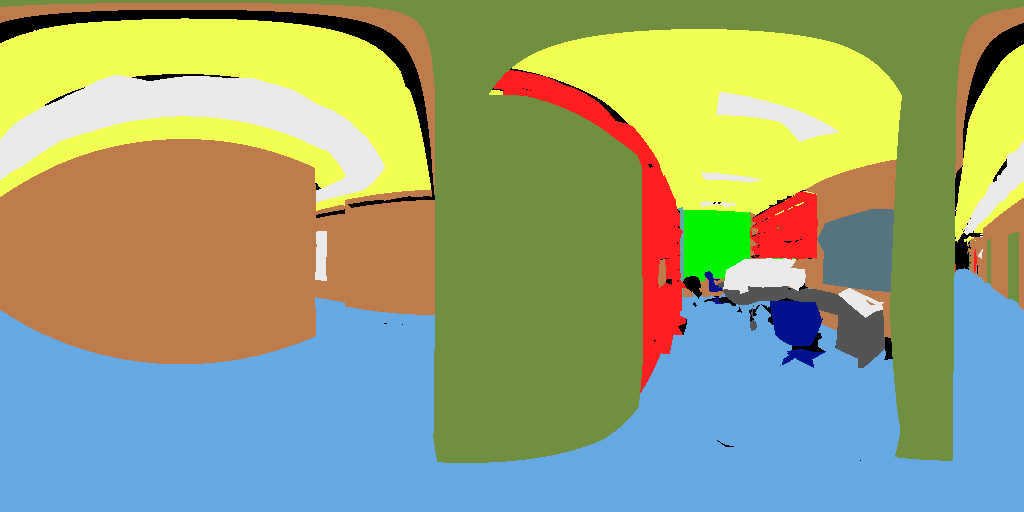}
        \end{subfigure}
    \end{subfigure}
    \begin{subfigure}{\textwidth}
        \centering
        \rotatebox[origin=l]{90}{\begin{minipage}{4pc}\caption*{~}\end{minipage}}%
        \begin{subfigure}{0.311\textwidth}
            \includegraphics[width=\textwidth]{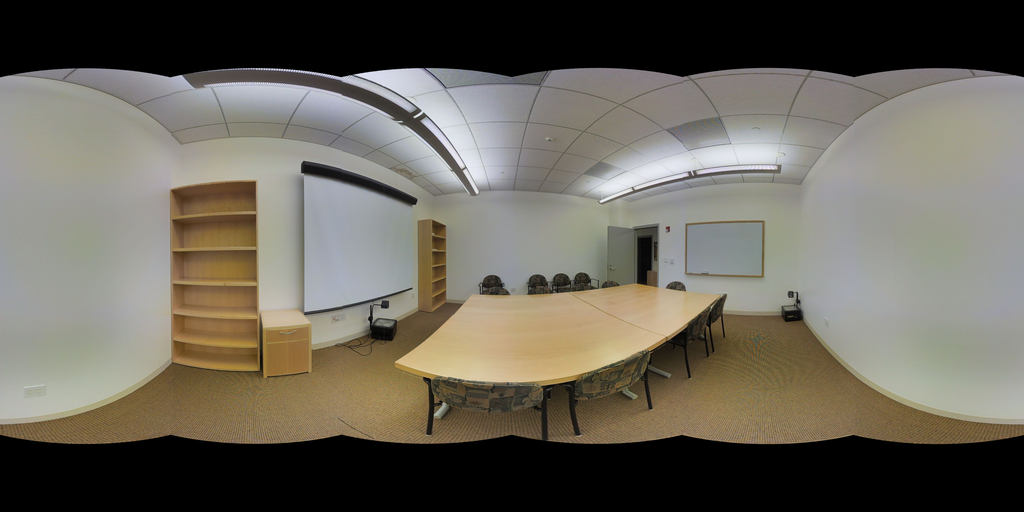}
            \vspace{-1.5em}
            \caption{\texttt{Scene 1}}
            \label{fig:scene_1}
        \end{subfigure}
        \begin{subfigure}{0.311\textwidth}
            \includegraphics[width=\textwidth]{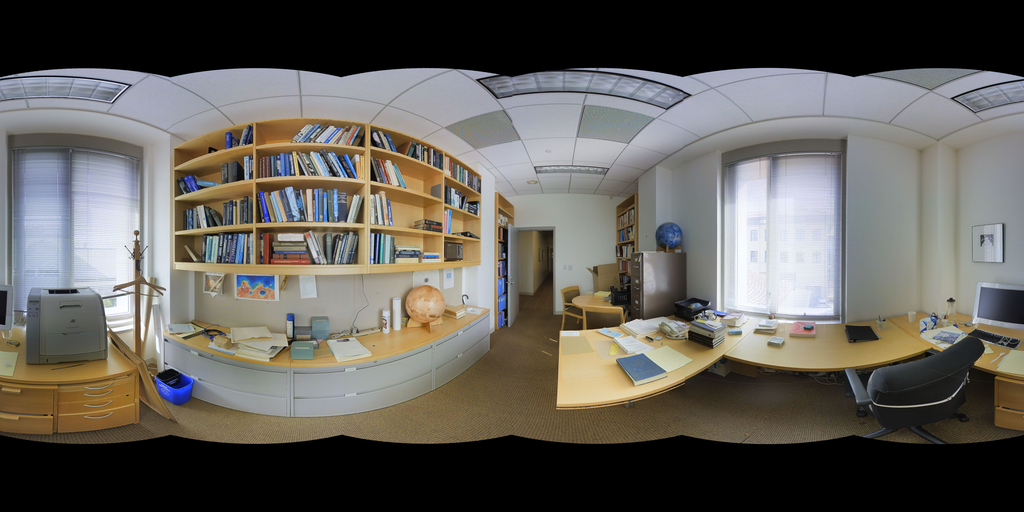}
            \vspace{-1.5em}
            \caption{\texttt{Scene 2}}
        \end{subfigure}
        \begin{subfigure}{0.311\textwidth}
            \includegraphics[width=\textwidth]{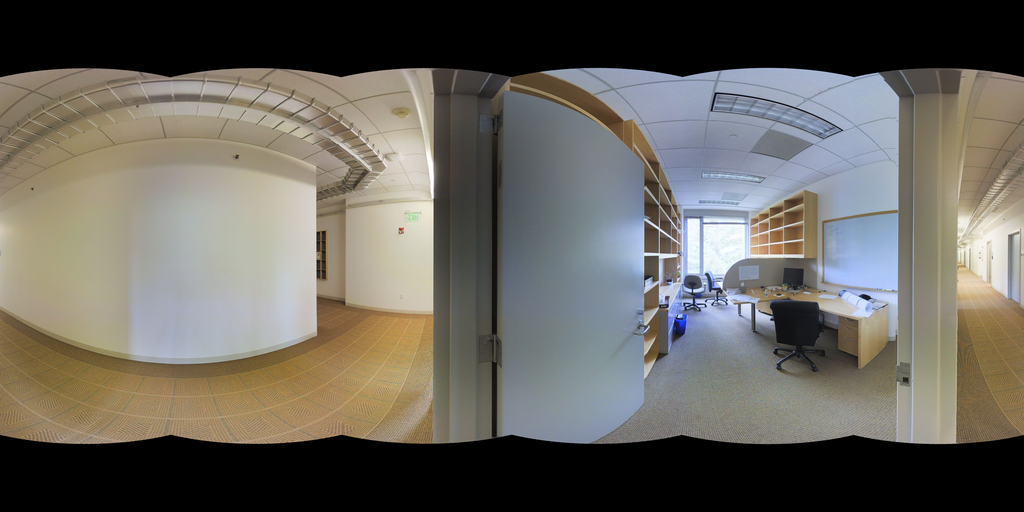}
            \vspace{-1.5em}
            \caption{\texttt{Scene 3}}
        \end{subfigure}
    \end{subfigure}
    \vspace{-2em}
    \caption{
        Comparison of the qualitative segmentation results of our PanoSAMic model for different scenes from the Stanford2D3DS dataset~\cite{armeni2017joint} and using different input configurations.
    }
    \label{fig:qualitative_results}
    \vspace{-1.5em}
\end{figure}


%% file: fig/fig_generalization.tex
\begin{figure}[t]
    \centering
    \begin{subfigure}{\textwidth}
        \centering
        \begin{subfigure}{0.32\textwidth}
            \includegraphics[width=\textwidth]{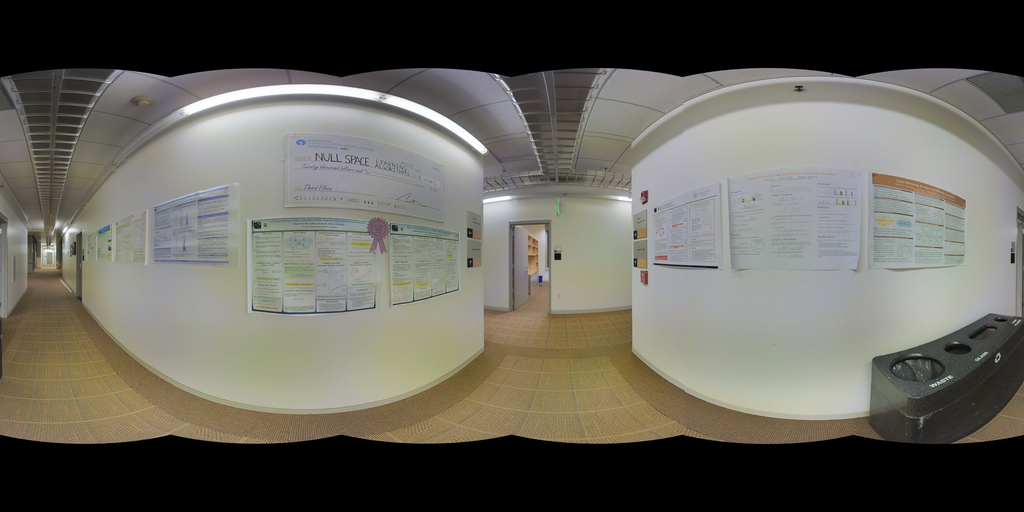}
        \end{subfigure}
        \begin{subfigure}{0.32\textwidth}
            \caption*{\textbf{\scriptsize{Ground Truth}}}
            \includegraphics[width=\textwidth]{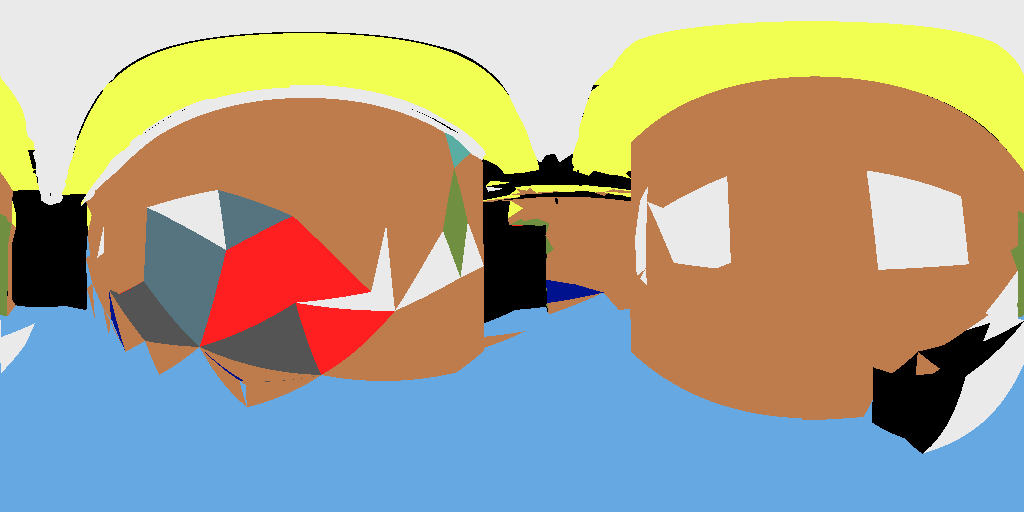}
        \end{subfigure}
        \begin{subfigure}{0.32\textwidth}
            \caption*{\textbf{\scriptsize{PanoSAMic (RGB-D-N)}}}
            \includegraphics[width=\textwidth]{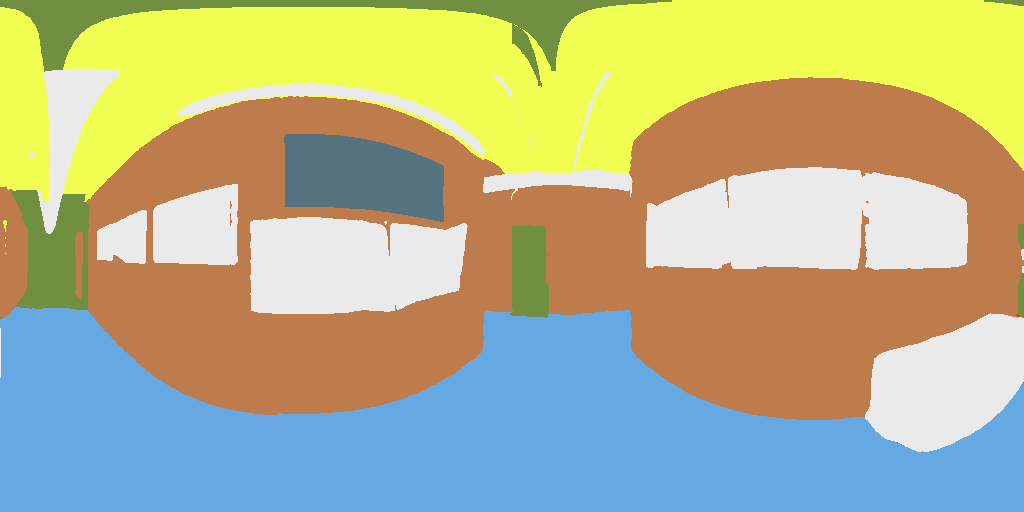}
        \end{subfigure}
        \vspace{-0.5em}
        \caption{Low ground truth imprecision}
        \label{subfig:low_dist}
    \end{subfigure}
    \begin{subfigure}{\textwidth}
        \centering
        \begin{subfigure}{0.32\textwidth}
            \includegraphics[width=\textwidth]{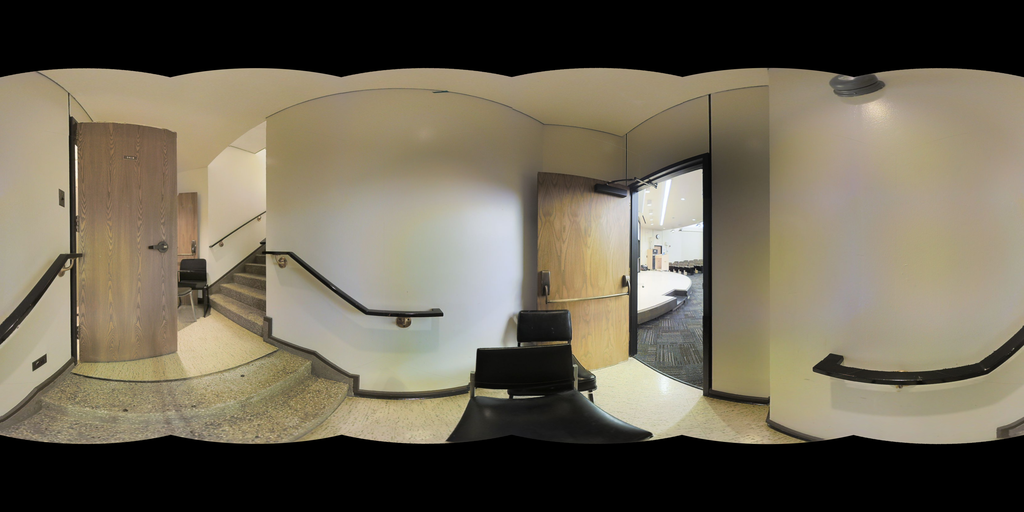}
        \end{subfigure}
        \begin{subfigure}{0.32\textwidth}
            \includegraphics[width=\textwidth]{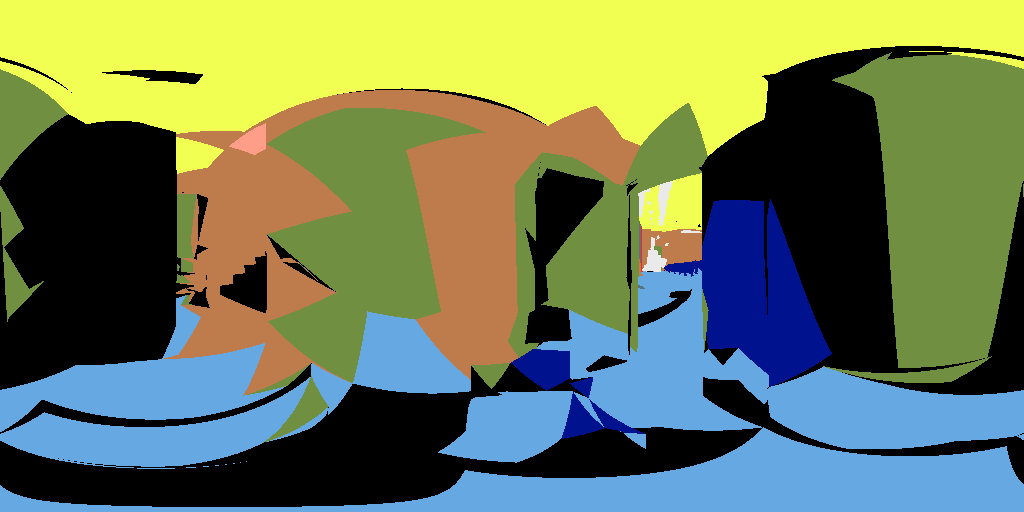}
        \end{subfigure}
        \begin{subfigure}{0.32\textwidth}
            \includegraphics[width=\textwidth]{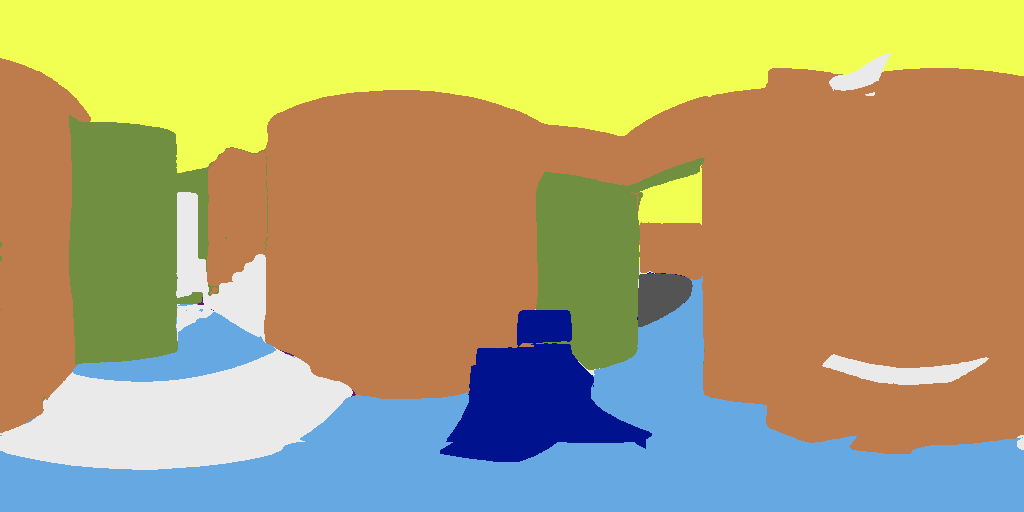}
        \end{subfigure}
        \vspace{-0.5em}
        \caption{Medium ground truth imprecision}
        \label{subfig:medium_dist}
    \end{subfigure}
    \begin{subfigure}{\textwidth}
        \centering
        \begin{subfigure}{0.32\textwidth}
            \includegraphics[width=\textwidth]{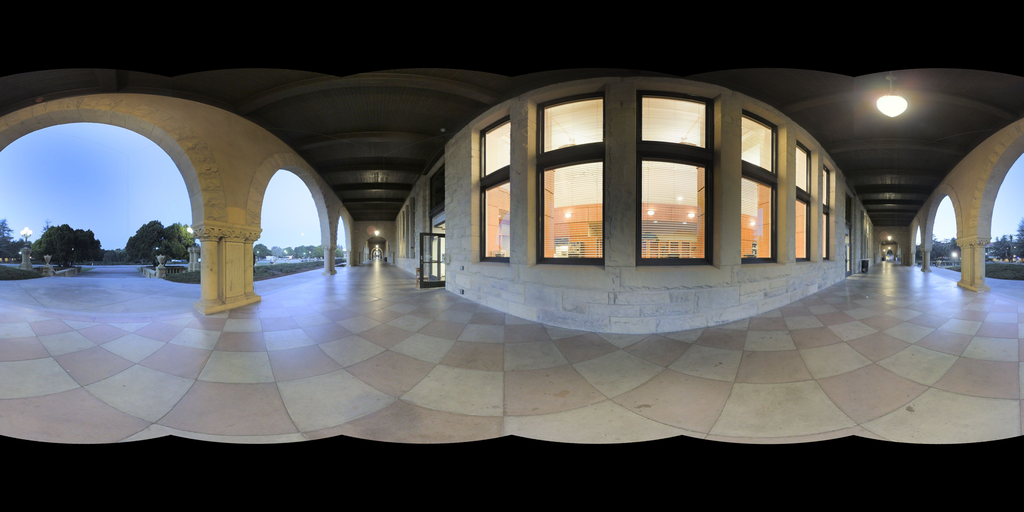}
        \end{subfigure}
        \begin{subfigure}{0.32\textwidth}
            \includegraphics[width=\textwidth]{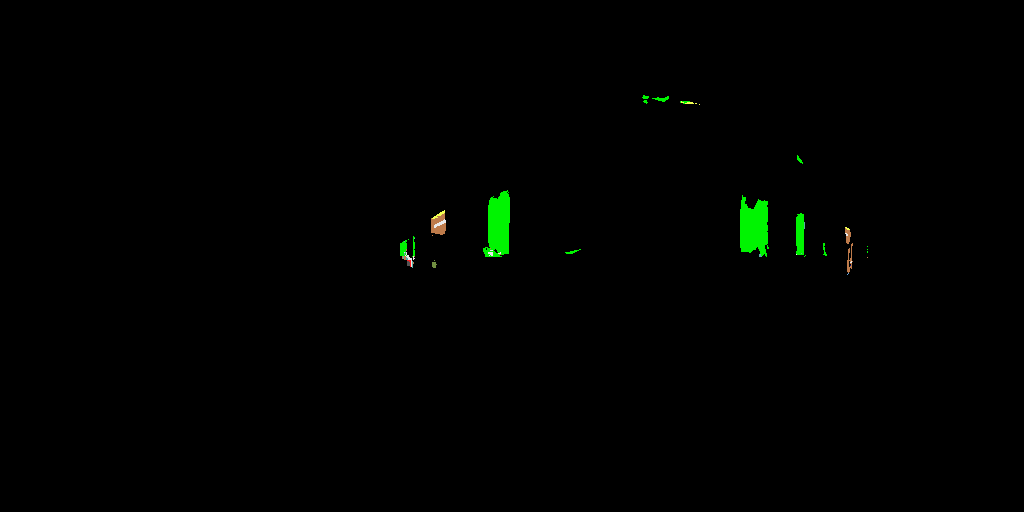}
        \end{subfigure}
        \begin{subfigure}{0.32\textwidth}
            \includegraphics[width=\textwidth]{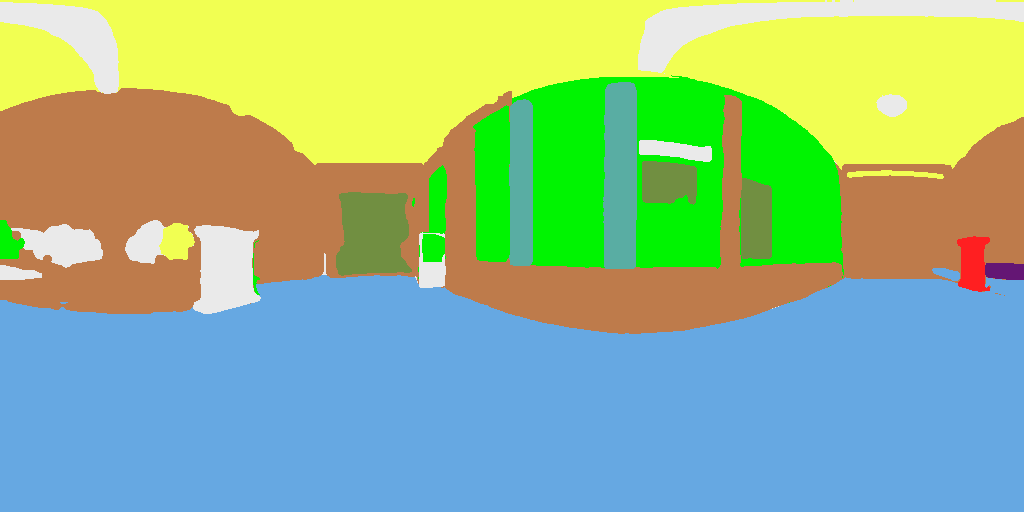}
        \end{subfigure}
        \vspace{-0.5em}
        \caption{High ground truth imprecision, missing annotations}
        \label{subfig:high_dist}
    \end{subfigure}
    \vspace{-2em}
    \caption{Some examples of scenes with imprecise ground truth labels from Stanford2D3DS~\cite{armeni2017joint}. Our PanoSAMic model performs very well on the different scenes and locations showing high generalization.}
    \label{fig:generalization}
    \vspace{-1.5em}
\end{figure}


%% file: tab/tab_ablation_study.tex
\begin{table}[b]
    \vspace{-2em}
    \centering
    \caption{Ablations of different model configurations and their effect on the segmentation results of Stanford2D3DS~\cite{armeni2017joint}.}
    \label{tab:ablation_study}
    \scriptsize
    \setlength{\tabcolsep}{6pt}
    \begin{tabular}{lcc} \toprule
        \textbf{Model Configuration}                                 & \textbf{mIoU $\%$} \\ \midrule \midrule
        \textbf{A:} (baseline) SAM Encoder + Conv. Decoder           & $53.29$ \\ \midrule
        \textbf{B:} A + Enc. Branches + \textit{no attention} (ours) & $61.20$ \\ \midrule
        \textbf{C:} B + Channel Attention (ours)                     & $60.82$ \\ \midrule
        \textbf{D:} B + CBAM (ours)                                  & $61.11$ \\ \midrule
        \textbf{E:} B + MCBAM (ours)                                 & $61.61$ \\ \midrule
        \textbf{F:} E + Instance Refinement (ours)                   & $\textbf{62.43}$ \\ \bottomrule
    \end{tabular}
    \vspace{-2em}
\end{table}

%% file: tab/tab_edge_ablation.tex
\begin{table}[t]
    \centering
    \caption{Comparison of the segmentation of the edges of input \textbf{RGB} between single-view and dual-view configurations.}
    \label{tab:edge_ablation}
    \scriptsize
    \setlength{\tabcolsep}{6pt}
    \begin{tabular}{ccccc} \toprule
        Edge  & \multicolumn{2}{c}{\textbf{Single View}}              & \multicolumn{2}{c}{\textbf{Dual View}} \\
        Ratio & \textbf{mIoU \%}                   & \textbf{mAcc \%} & \textbf{mIoU \%}                   & \textbf{mAcc \%} \\ \midrule \midrule
        -     & $57.95$                            & $71.79$          & $59.62$                            & $74.11$\\ \midrule
        $0.5$ & $57.79\textcolor{red}{\downarrow}$ & $72.14$          & $59.86\textcolor{green}{\uparrow}$ & $74.83$\\ \midrule
        $0.3$ & $57.51\textcolor{red}{\downarrow}$ & $72.17$          & $60.23\textcolor{green}{\uparrow}$ & $75.56$\\ \midrule
        $0.1$ & $55.52\textcolor{red}{\downarrow}$ & $70.54$          & $60.39\textcolor{green}{\uparrow}$ & $76.21$\\ \bottomrule
    \end{tabular}
    \vspace{-1em}
\end{table}

%% file: sec/6_conclusion.tex
\section{Conclusion} \label{sec:conclusion}

In this work, we presented \textit{PanoSAMic}, our multi-modal panoramic segmentation model using the SAM feature encoder and leveraging its pre-trained weights on large amounts of data. We extended the existing SAM architecture with our dual-view fusion to handle edge discontinuity of spherical images and introduced an improved self attention block (MCBAM) for multi-modal fusion and segmentation.

We evaluated our model on public data and achieved State-of-the-Art results by a large margin. We tested our model with different modalities including depth and normals and show that we achieve SotA with all input combinations. We further validated the importance of our architecture contributions through multiple ablations and proved that our Moving CBAM refines features for semantic segmentation tasks unlike the original CBAM module tailored for classification. We also showed that our dual view fusion successfully addresses the edge discontinuity of panoramic images. Our results demonstrate strong generalization capabilities and show that SAM can be adapted for semantic segmentation.

\section*{Acknowledgements}

This research was funded by the European Union as part of the projects: HumanTech (Grant Agreement 101058236) and ShieldBOT (Grant Agreement 101235093).

%% file: sec/X_suppl.tex
\clearpage
\setcounter{page}{1}
\title{
    PanoSAMic: Panoramic Image Segmentation from SAM Feature Encoding and Dual View Fusion
}
\subtitle{
    Supplementary Material
}
\titlerunning{PanoSAMic}

\author{
    Mahdi Chamseddine\inst{1,2} \and
    Didier Stricker\inst{1,2} \and
    Jason Rambach\inst{1}
}
\authorrunning{M. Chamseddine et al.}
\institute{
    German Research Center for Artificial Intelligence (DFKI), Kaiserslautern, Germany \and
    RPTU Kaiserslautern-Landau, Kaiserslautern, Germany \\
    \email{firstname.lastname@dfki.de}
}

\renewcommand{\thesection}{\Alph{section}}

\maketitle

\setcounter{figure}{6}
\setcounter{table}{4}

\section{Instance Guided Refinement} \label{sec:supp_refinement}

PanoSAMic extends SAM's~\cite{kirillov2023segment} instance segmentation capabilities to multi-modal panoramic scenes. The model produces two complementary outputs: (1) instance segmentation masks from SAM's frozen components, and (2) semantic segmentation logits from trainable fusion and decoder modules. Instance generation uses SAM's automatic mask generator with a $32\times32$ grid of point prompts, processed in batches of $64$.

\subsection{Multi-Modal Instance Generation}

\subsubsection{Modality-wise segmentation:} Unlike SAM which processes only RGB, we generate instance masks from all available modalities (RGB, depth, normals). For each modality, we independently apply SAM's prompt encoder and mask decoder, producing separate sets of instance proposals. This multi-modal approach captures complementary boundaries: RGB excels at texture edges, depth captures geometric discontinuities, and normals detect surface orientation changes.

\subsubsection{Post-processing:} Each modality's predictions undergo SAM's standard post-processing pipeline: stability score filtering, predicted IoU filtering, box-based NMS (non-maximum suppression), and small region removal.

\subsubsection{Cross-modality fusion:} After individual modality processing, we merge masks from all modalities using greedy mask-based NMS. Masks are sorted by quality score (predicted IoU), and lower-quality masks overlapping with higher-quality ones are removed. This preserves the best boundaries from each modality.

\subsection{Dual-View Panoramic Fusion}

Masks from the rotated view are \textit{unshifted} back to original coordinates. When masks from both views overlap significantly, we select the higher-quality mask based on predicted IoU. If quality scores differ by a small margin, we use mask area as a tiebreaker, preferring larger masks. This ensures we keep the best representation of each object regardless of which view captured it better.

\subsection{Semantic Refinement}

Given semantic logits $\mathbf{L} \in \mathbb{R}^{C \times H \times W}$ from the decoder and filtered instance masks, we refine predictions as follows: for each instance mask, we compute the most frequent semantic class within that instance region (using initial argmax predictions), then assign all pixels in that instance to this majority class. Background pixels (not covered by any instance) retain their original semantic predictions.

\subsection{Evaluation}

\Cref{fig:refinement} shows the effect of refinement on the prediction of the different modality inputs. The refinement step improves the overall quality of the segmentation, reduces ``blobiness'', and enhances the edges. In rare cases, the refinement can have a negative effect by removing the correct class if it is not well segmented in the data: \eg~the column in the RGB example, or the clutter on the bookshelf in the RGB-D and RGB-D-N examples.

\input{fig/fig_refinement}

\section{SAM3 Evaluation} \label{sec:supp_sam3}

SAM3~\cite{carion2025sam} is a text-promptable segmentation model that generates instance masks conditioned on natural language class descriptions. We used the official SAM3 checkpoint and code and evaluation is performed on all three folds of Stanford2D3DS and the single validation split of Matterport3D and results reported in~\Cref{tab:stanford2d3ds,tab:matterport3d}.

\subsection{Evaluation Procedure}

For each test image, we perform per-class prompting by sequentially querying SAM3 with text prompts corresponding to each semantic class name (\eg, ``wall'', ``floor'', ``ceiling''). Each prompt generates a set of scored instance masks. We fuse these per-class predictions into a unified semantic segmentation map by constructing a class score tensor of shape $C \times H \times W$. For each class $c$ at pixel $(x,y)$, we compute:
\begin{equation}
    s_c(x,y) = \max_{i} m_i(x,y) \cdot \sigma_i,
\end{equation}
where $m_i$ is the mask logit from prediction $i$ for class $c$, and $\sigma_i$ is its confidence score. Only masks with confidence $\sigma_i \geq 0.25$ are retained. The final semantic label assigns each pixel to the class with the highest score.

\subsection{Clutter Class Handling}
 
Both datasets include a catch-all class for miscellaneous objects: ``clutter'' in Stanford2D3DS and ``objects'' in Matterport3D. For pixels with no coverage (all class scores are zero), we assign them to the clutter class. Additionally, pixels where the maximum class score falls below $0.05$ are also assigned to clutter. This effectively creates a low-confidence region classifier.

\subsection{Spatial Smoothing}

To reduce speckle artifacts, we apply $3 \times 3$ majority filtering after obtaining the initial label predictions. For each pixel, we replace its label with the most frequent label in its connected neighborhood. This smoothing is applied after clutter assignment to ensure it operates on the final label space including the clutter class.

\section{Evaluation of Encoder Size} \label{sec:supp_encoders}

\Cref{tab:model_size} shows the results of testing the different encoder sizes pretrained by SAM~\cite{kirillov2023segment} on the Stanford2D3DS~\cite{armeni2017joint} dataset. While the ViT-L encoder performs slightly worse than ViT-H, the ViT-B encoder shows significant result degradation.

\input{tab/tab_model_size}

Overall, our model still delivers competitive even state-of-the-art semantic segmentation results regardless of the SAM encoder used.

\section{Model Parameters and Efficiency} \label{sec:supp_params}

Our model uses the frozen SAM encoder for its backbone. The number of parameters for SAM are as reported in their paper~\cite{kirillov2023segment}: 91M, 308M, and 636M parameters for the ViT-B, ViT-L, and ViT-H backbones respectively. Our trainable parameters and model FLOPs shown in~\Cref{tab:model_params} for the different modality inputs and lower and upper bounds using the ViT-B and ViT-H encoders.

\input{tab/tab_model_params}

\section{More Qualitative Results} \label{sec:supp_results}
\Cref{fig:more_qualitative_results} shows some more qualitative results of our \textbf{RGB-D-N} model configuration. The results agree with our reported quantitative results in~\Cref{tab:stanford2d3ds}.

The comparison also shows that our model even surpasses the ground truth in some places: pillow on a sofa is classified as clutter, better detection of label edges, reliable class prediction for missing ground truth areas, etc.

\input{fig/fig_more_qualitative_results}

%% file: fig/fig_refinement.tex
\begin{figure}[t]
    \centering
    \begin{subfigure}{\textwidth}
        \centering
        \rotatebox[origin=l]{90}{\begin{minipage}{4pc}\caption*{\textbf{\scriptsize{without}}}\end{minipage}}%
        \begin{subfigure}{0.311\textwidth}
            \caption*{\textbf{\scriptsize{RGB}}}
            \includegraphics[width=\textwidth]{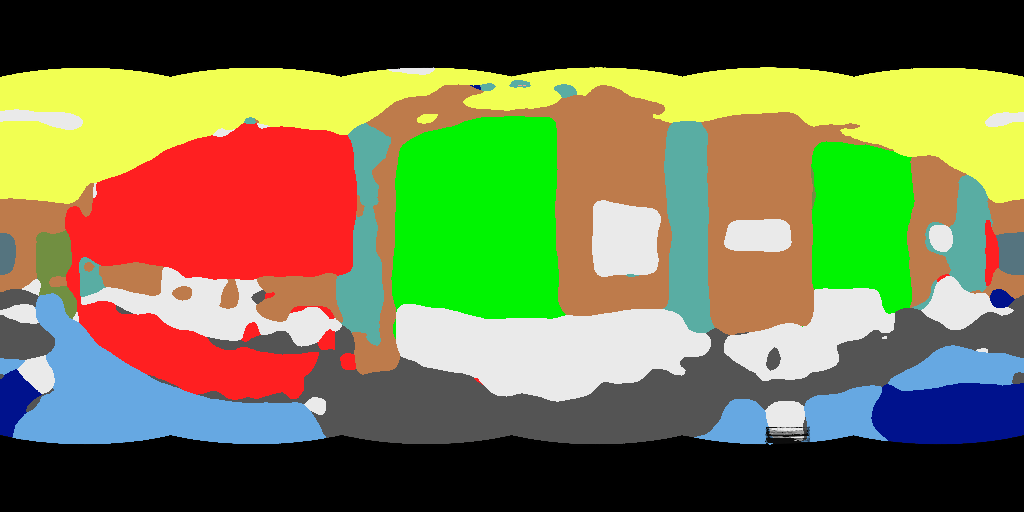}
        \end{subfigure}
        \begin{subfigure}{0.311\textwidth}
            \caption*{\textbf{\scriptsize{RGB-D}}}
            \includegraphics[width=\textwidth]{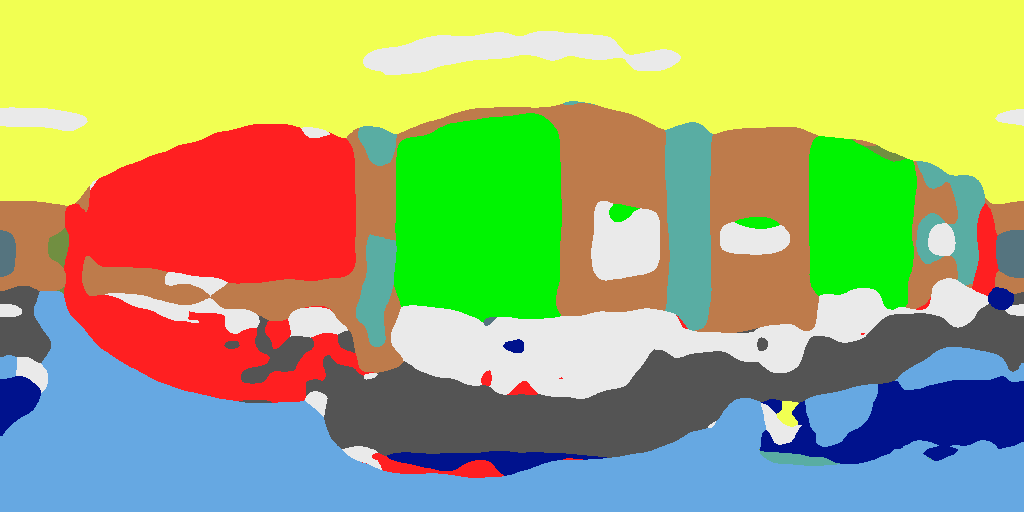}
        \end{subfigure}
        \begin{subfigure}{0.311\textwidth}
            \caption*{\textbf{\scriptsize{RGB-D-N}}}
            \includegraphics[width=\textwidth]{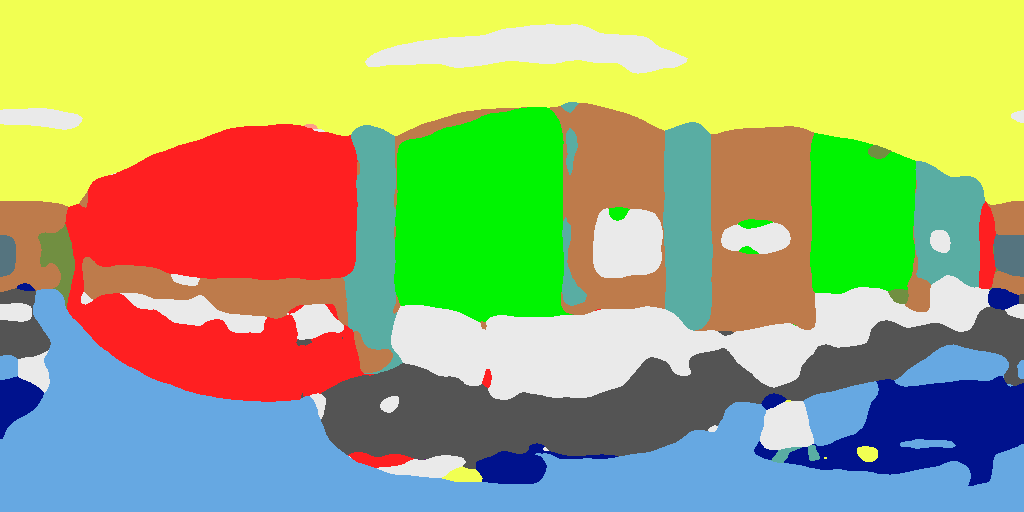}
        \end{subfigure}
    \end{subfigure}
    \begin{subfigure}{\textwidth}
        \centering
        \rotatebox[origin=l]{90}{\begin{minipage}{4pc}\caption*{\textbf{\scriptsize{with}}}\end{minipage}}%
        \begin{subfigure}{0.311\textwidth}
            \includegraphics[width=\textwidth]{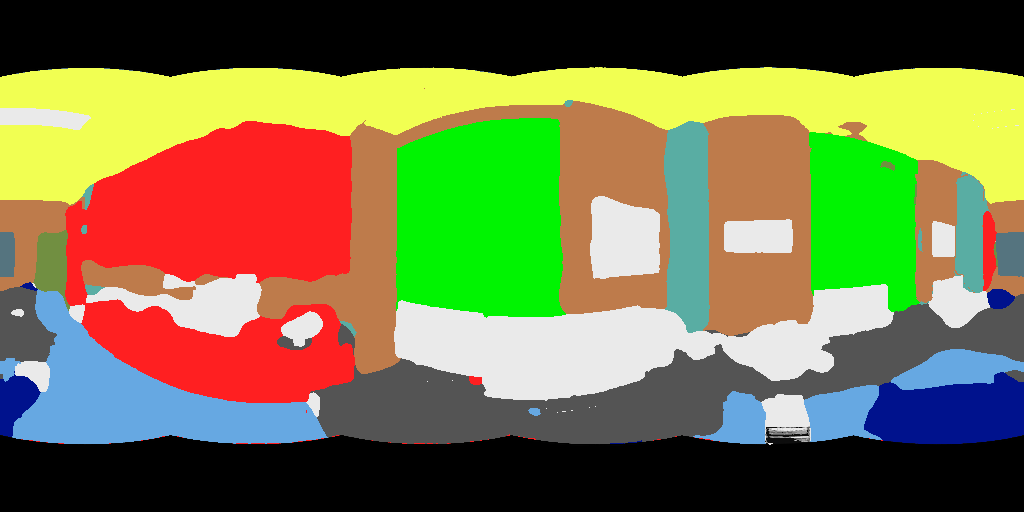}
        \end{subfigure}
        \begin{subfigure}{0.311\textwidth}
            \includegraphics[width=\textwidth]{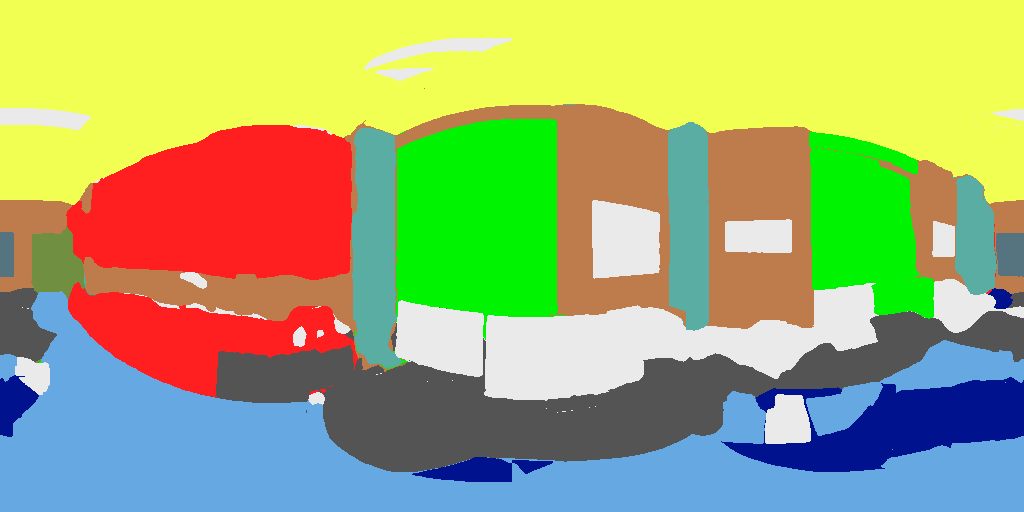}
        \end{subfigure}
        \begin{subfigure}{0.311\textwidth}
            \includegraphics[width=\textwidth]{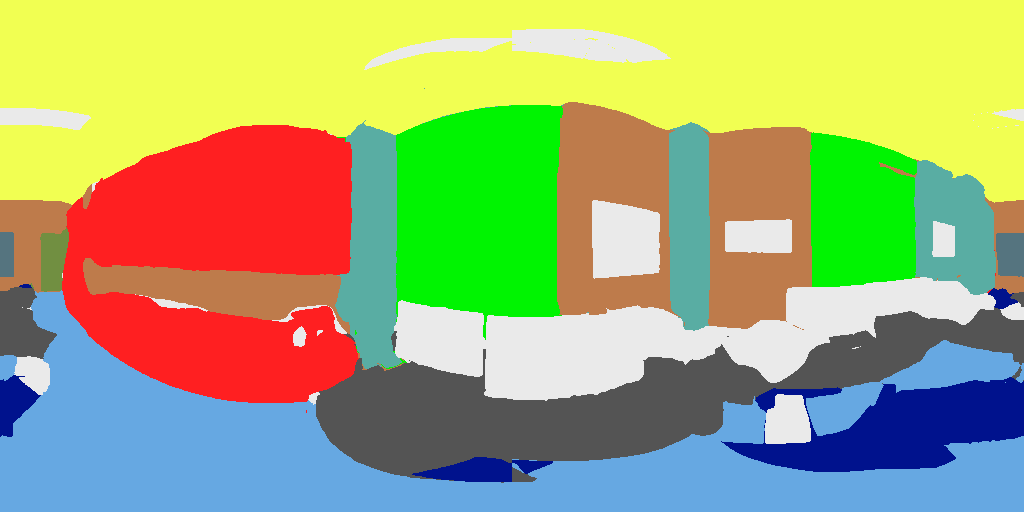}
        \end{subfigure}
    \end{subfigure}
    \begin{subfigure}{\textwidth}
        \centering
        \rotatebox[origin=l]{90}{\begin{minipage}{4pc}\caption*{\textbf{\scriptsize{GT}}}\end{minipage}}%
        \begin{subfigure}{0.311\textwidth}
            \includegraphics[width=\textwidth]{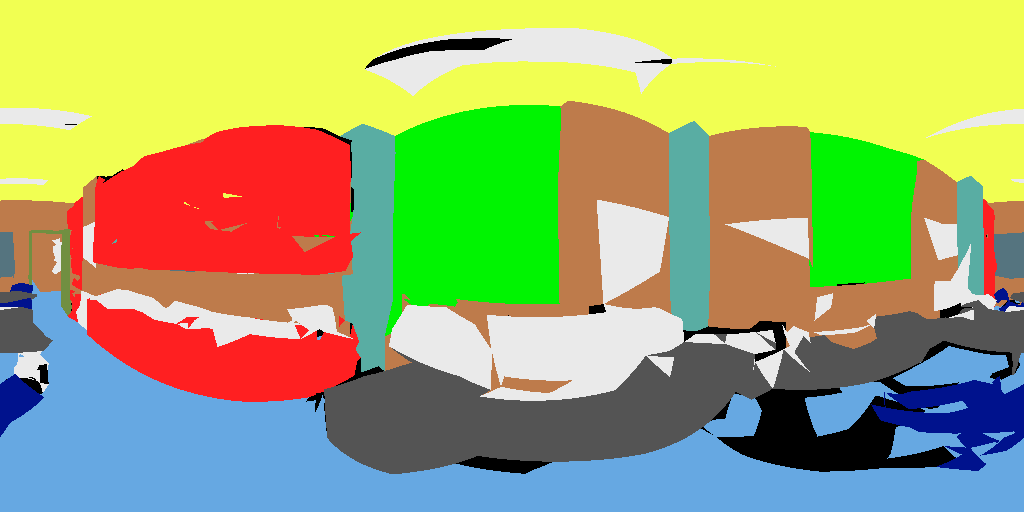}
            \vspace{-1.5em}
            \caption*{\textbf{\scriptsize{Ground Truth}}}
        \end{subfigure}
        \begin{subfigure}{0.311\textwidth}
            \includegraphics[width=\textwidth]{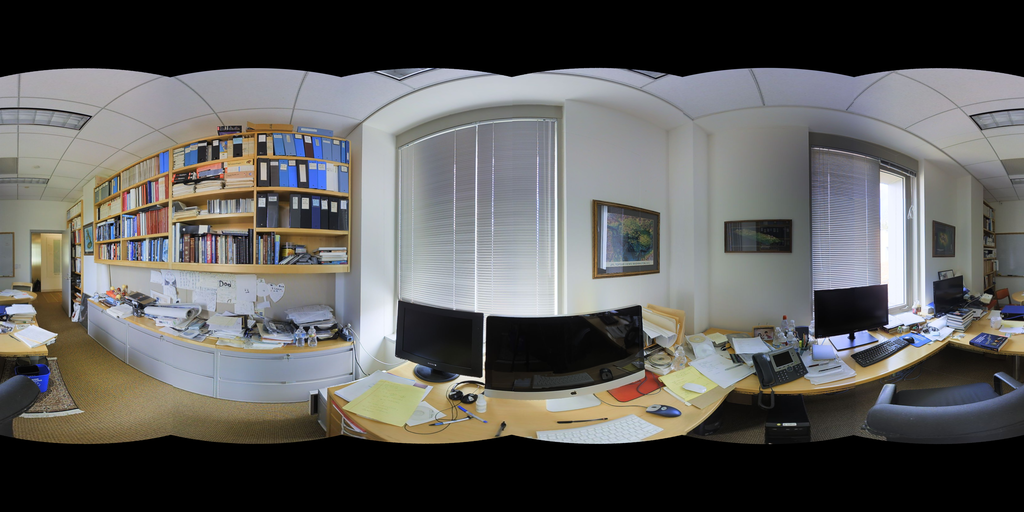}
            \vspace{-1.5em}
            \caption*{}
        \end{subfigure}
    \end{subfigure}
    \vspace{-2em}
    \caption{
        Comparison of the qualitative segmentation results before and after refinement on different modality inputs.
    }
    \label{fig:refinement}
    \vspace{-1.5em}
\end{figure}


%% file: tab/tab_model_size.tex
\begin{table}
    \centering
    \caption{Comparing the segmentation results (3-Fold) with respect to different encoder depths.}
    \label{tab:model_size}
    \scriptsize
    \setlength{\tabcolsep}{6pt}
    \begin{tabular}{cccc} \toprule
        ~              & \textbf{Encoder Depth} & \textbf{mIoU \%} & \textbf{mAcc \%} \\ \midrule \midrule
        
        \textbf{ViT-B} & $12$                   & $56.68$          & $70.49$ \\
        \textbf{ViT-L} & $24$                   & $60.90$          & $73.09$ \\
        \textbf{ViT-H} & $32$                   & $61.57$          & $74.04$ \\ \bottomrule
    \end{tabular}
\end{table}

%% file: tab/tab_model_params.tex
\begin{table}
    \vspace{-3em}
    \centering
    \caption{Trainable parameters and full model FLOPs.}
    \label{tab:model_params}
    \scriptsize
    \setlength{\tabcolsep}{6pt}
    \begin{tabular}{lcc} \toprule
        \textbf{Modalities} & \textbf{\# M. Params} & \textbf{TFLOPS} \\ \midrule
        
        RGB     & $139-178$ & $1.4-6.3$ \\
        RGB-D   & $141-184$ & $2.7-12.7$ \\
        RGB-D-N & $144-191$ & $4.1-19.0$ \\ \bottomrule
    \end{tabular}
    \vspace{-4em}
\end{table}

%% file: fig/fig_more_qualitative_results.tex
\begin{figure}[ht]
    \centering
    \begin{subfigure}{\textwidth}
        \begin{subfigure}{0.32\textwidth}
            \includegraphics[width=\textwidth]{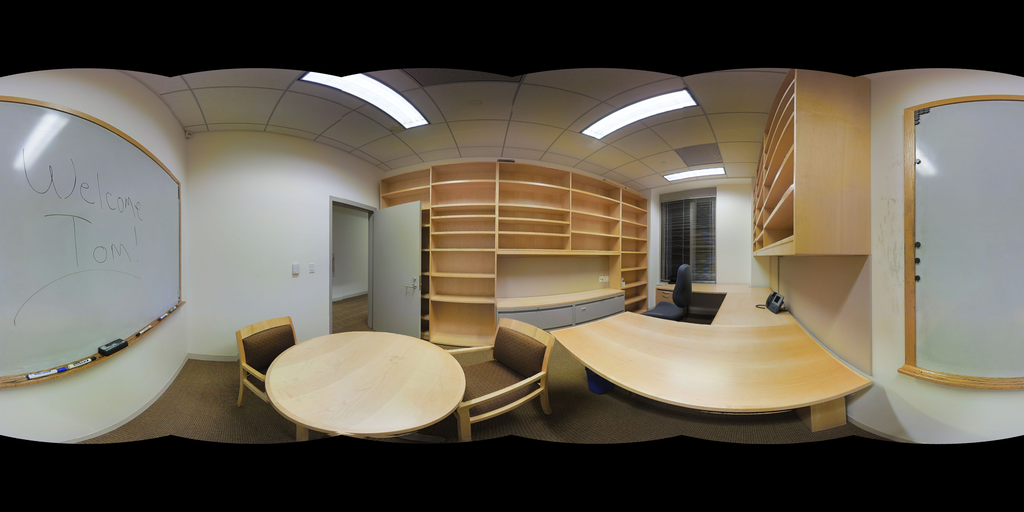}
        \end{subfigure}
        \begin{subfigure}{0.32\textwidth}
            \caption*{\textbf{\scriptsize{Ground Truth}}}
            \includegraphics[width=\textwidth]{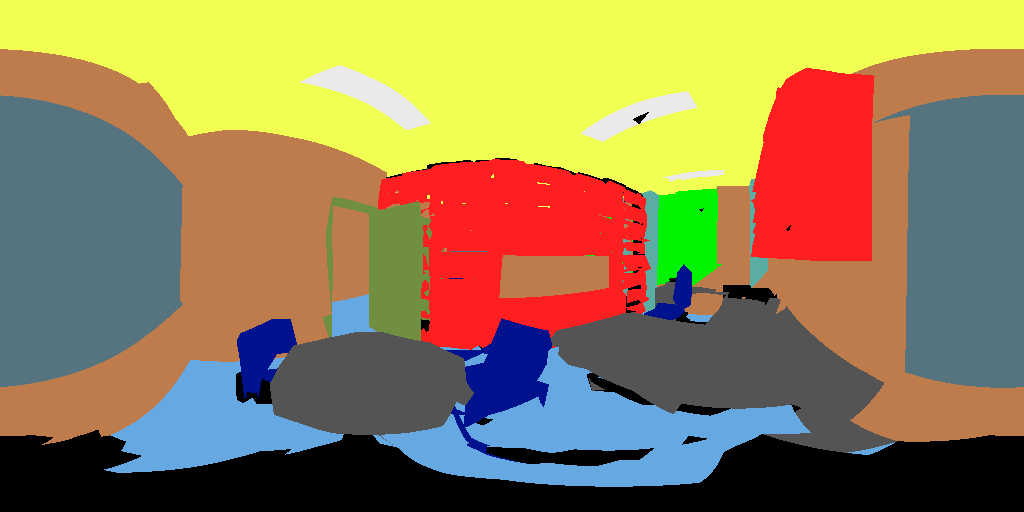}
        \end{subfigure}
        \begin{subfigure}{0.32\textwidth}
            \caption*{\textbf{\scriptsize{PanoSAMic (RGB-D-N)}}}
            \includegraphics[width=\textwidth]{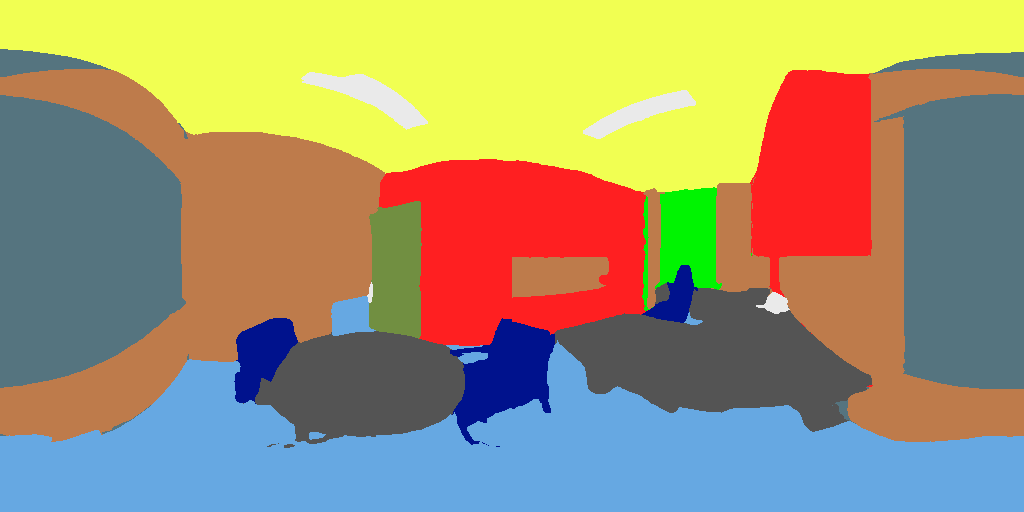}
        \end{subfigure}
    \end{subfigure}
    \begin{subfigure}{\textwidth}
        \begin{subfigure}{0.32\textwidth}
            \includegraphics[width=\textwidth]{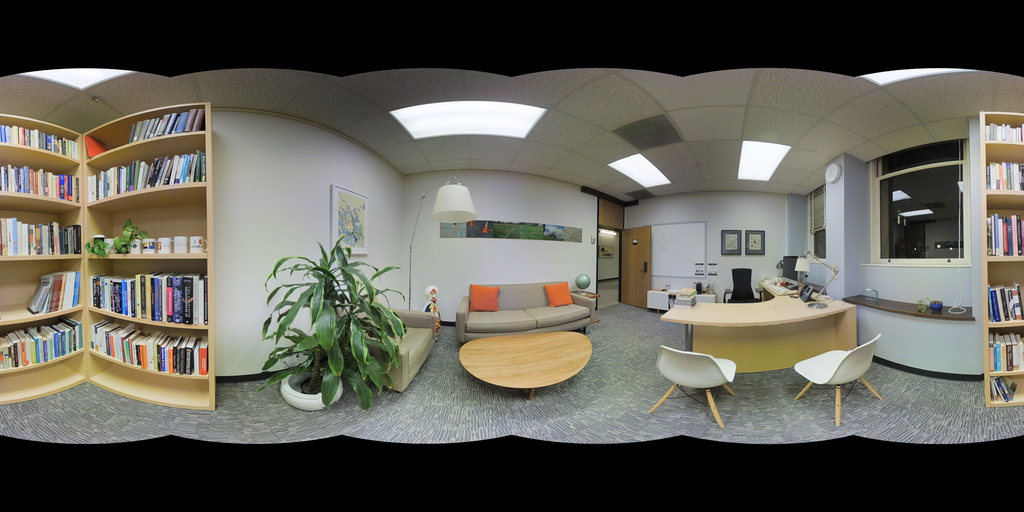}
        \end{subfigure}
        \begin{subfigure}{0.32\textwidth}
            \includegraphics[width=\textwidth]{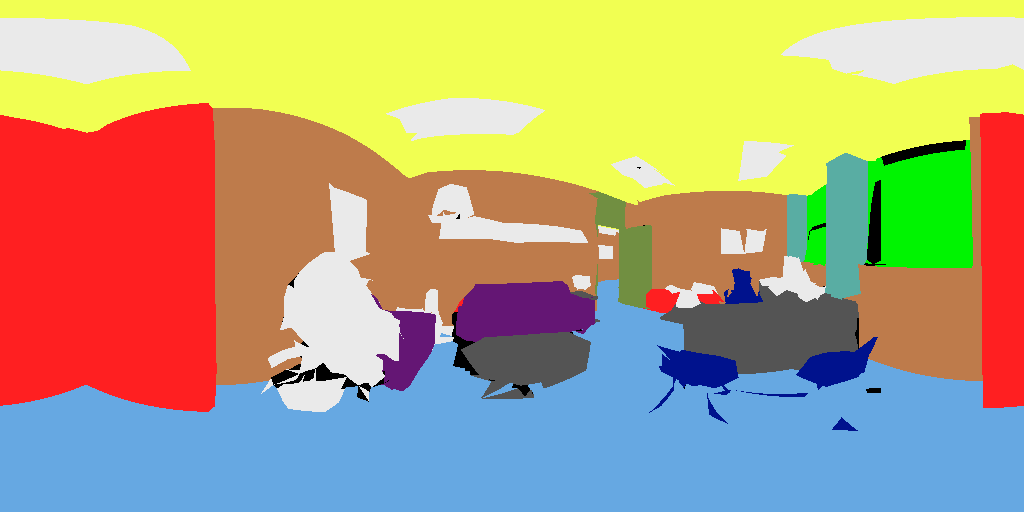}
        \end{subfigure}
        \begin{subfigure}{0.32\textwidth}
            \includegraphics[width=\textwidth]{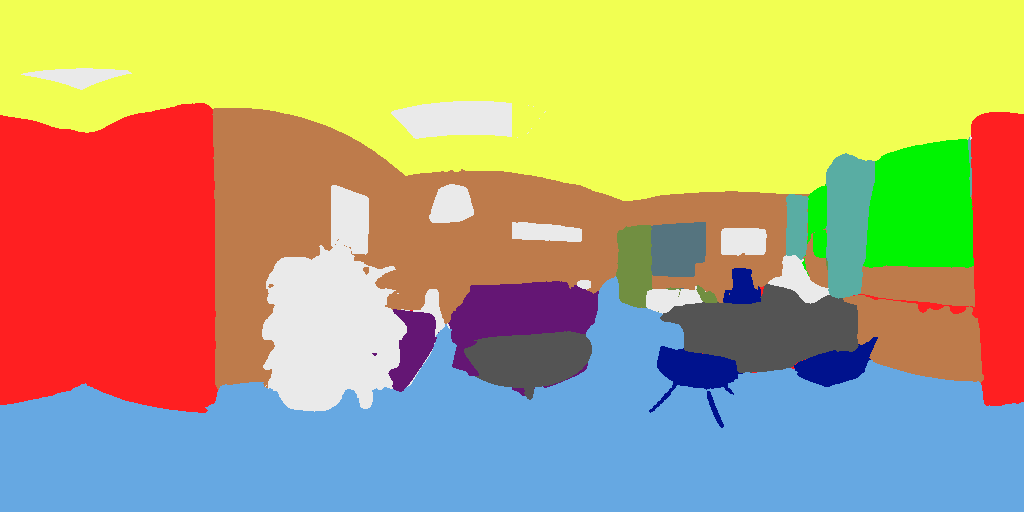}
        \end{subfigure}
    \end{subfigure}
    \begin{subfigure}{\textwidth}
        \begin{subfigure}{0.32\textwidth}
            \includegraphics[width=\textwidth]{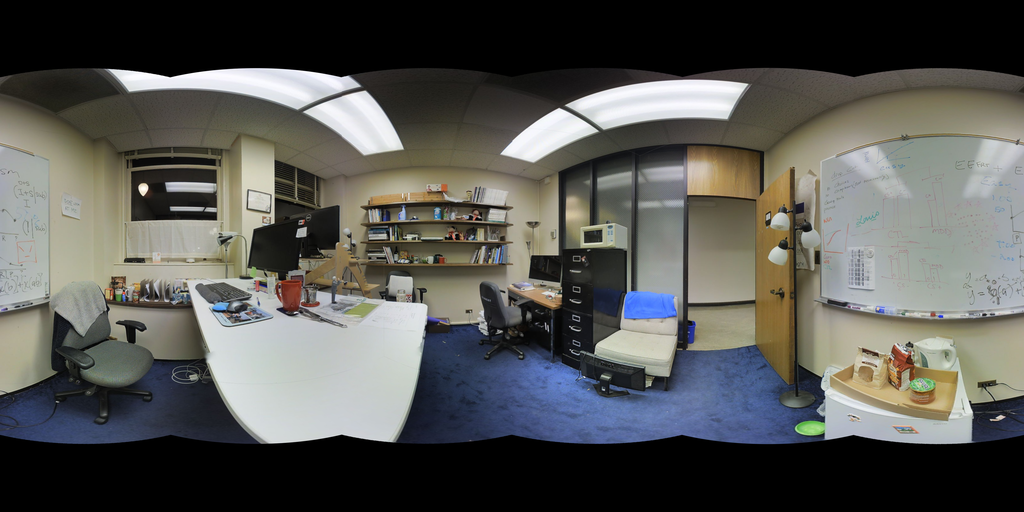}
        \end{subfigure}
        \begin{subfigure}{0.32\textwidth}
            \includegraphics[width=\textwidth]{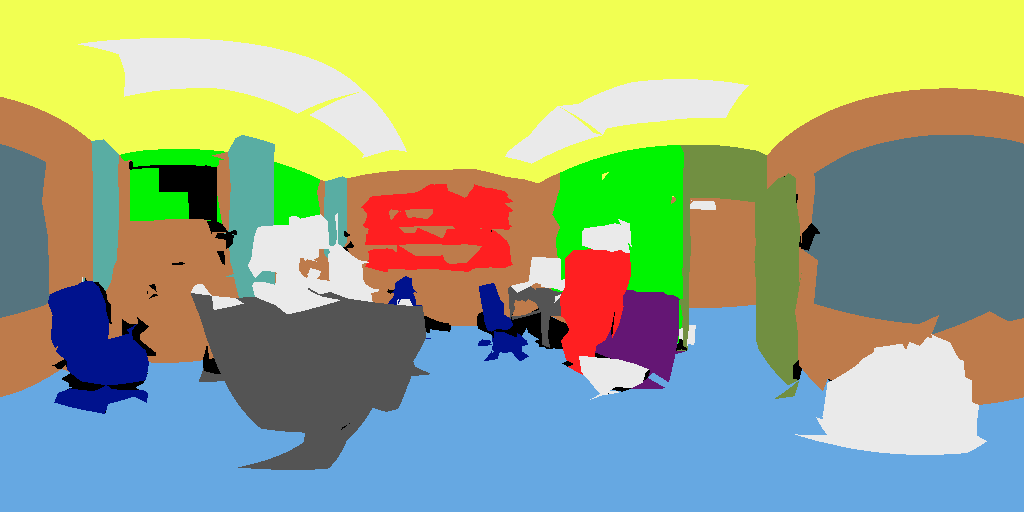}
        \end{subfigure}
        \begin{subfigure}{0.32\textwidth}
            \includegraphics[width=\textwidth]{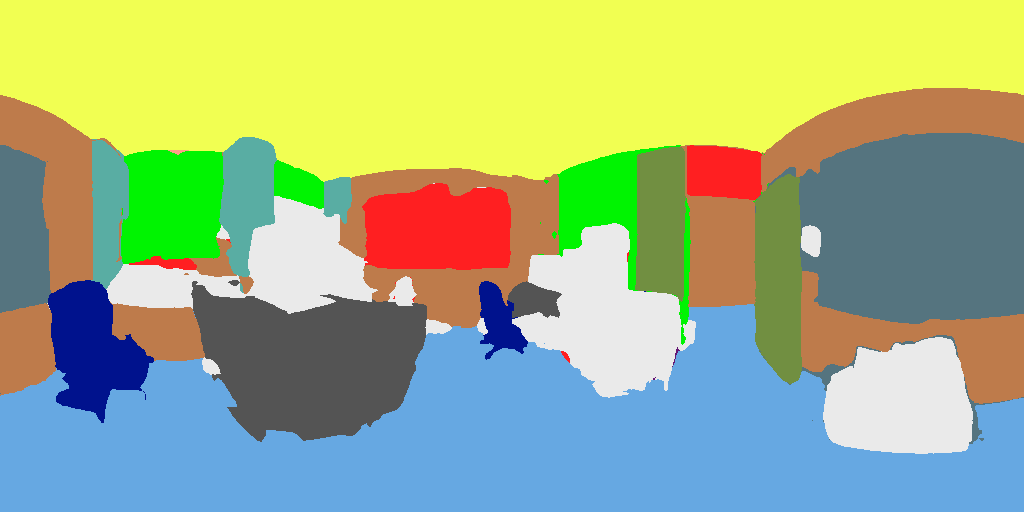}
        \end{subfigure}
    \end{subfigure}
    \begin{subfigure}{\textwidth}
        \begin{subfigure}{0.32\textwidth}
            \includegraphics[width=\textwidth]{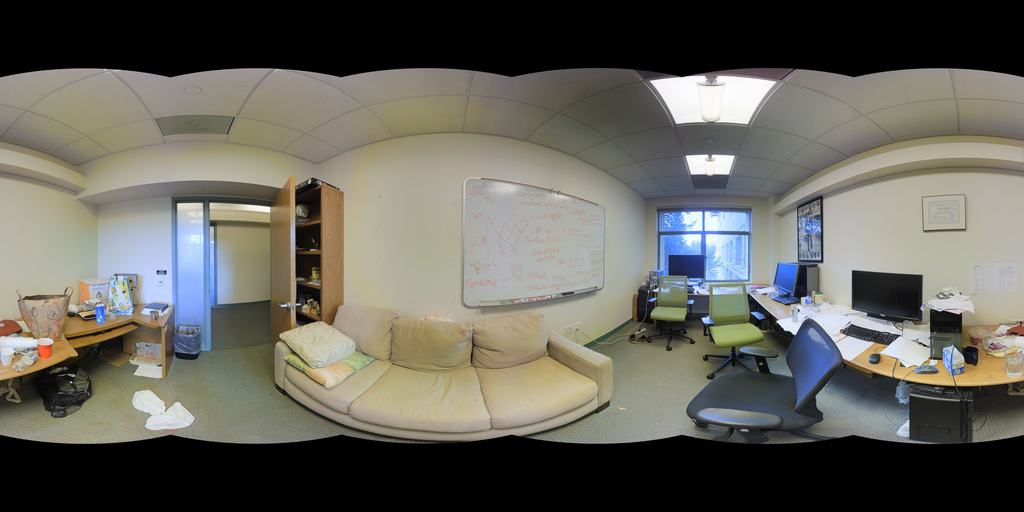}
        \end{subfigure}
        \begin{subfigure}{0.32\textwidth}
            \includegraphics[width=\textwidth]{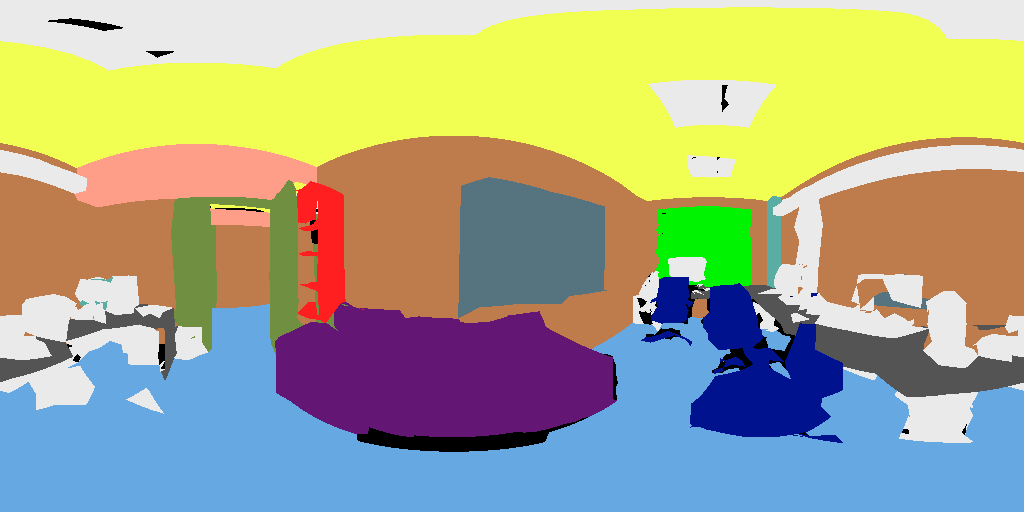}
        \end{subfigure}
        \begin{subfigure}{0.32\textwidth}
            \includegraphics[width=\textwidth]{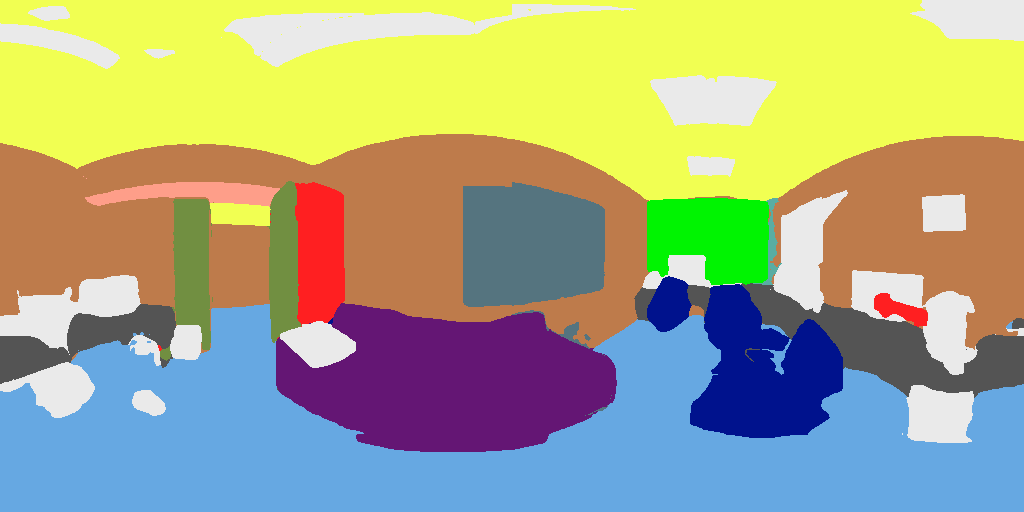}
        \end{subfigure}
    \end{subfigure}
    \begin{subfigure}{\textwidth}
        \begin{subfigure}{0.32\textwidth}
            \includegraphics[width=\textwidth]{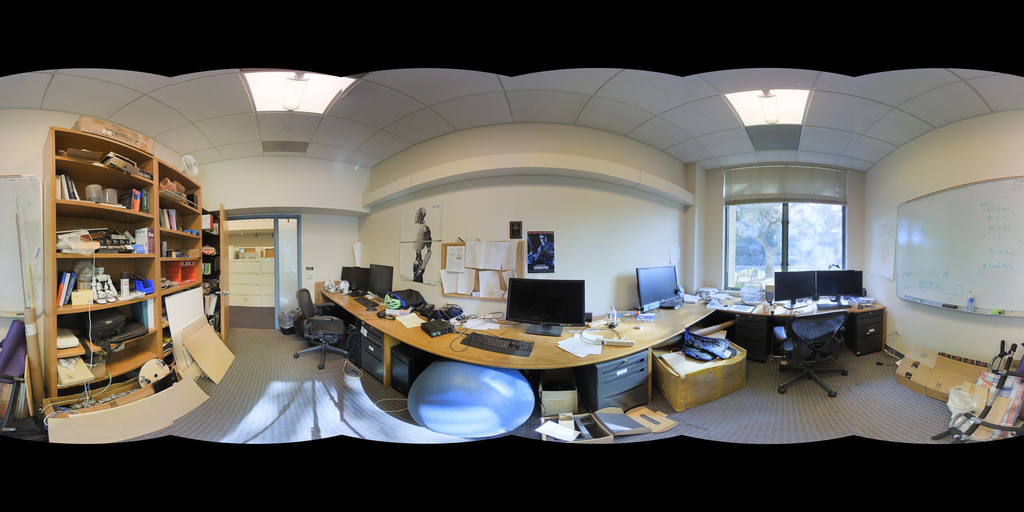}
        \end{subfigure}
        \begin{subfigure}{0.32\textwidth}
            \includegraphics[width=\textwidth]{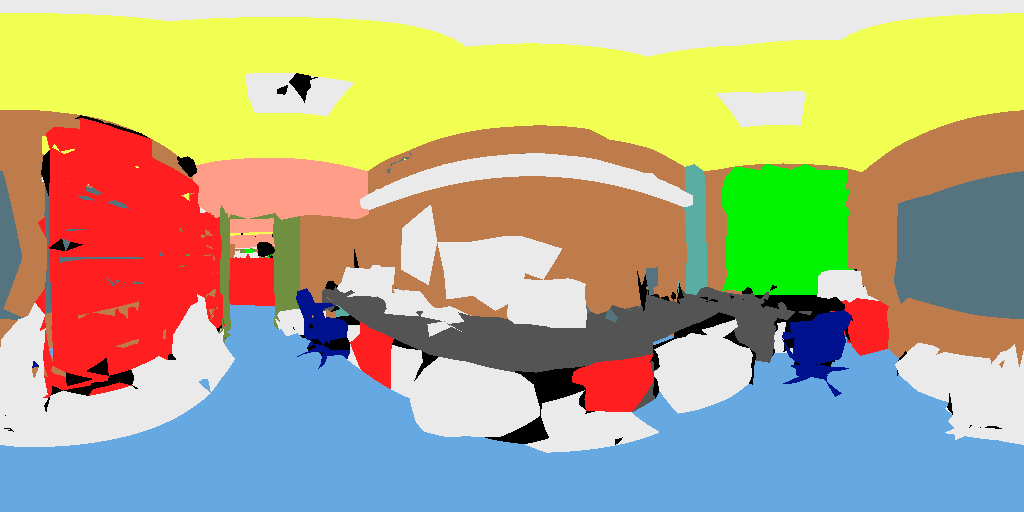}
        \end{subfigure}
        \begin{subfigure}{0.32\textwidth}
            \includegraphics[width=\textwidth]{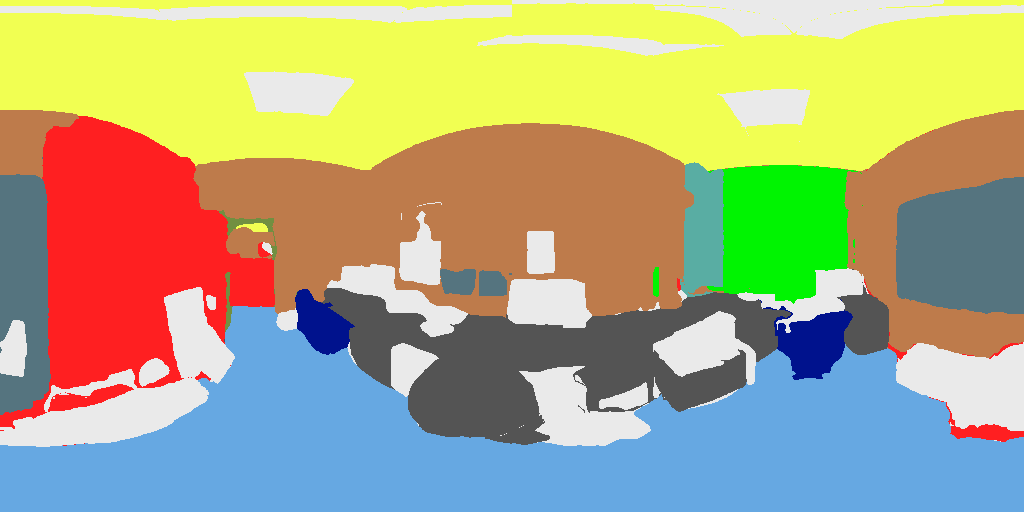}
        \end{subfigure}
    \end{subfigure}
    \vspace{-2em}
    \caption{More qualitative results on the Stanford2D3DS dataset~\cite{armeni2017joint}.}
    \label{fig:more_qualitative_results}
    \vspace{-1em}
\end{figure}
